\documentclass[10pt,twocolumn,letterpaper]{article}

\usepackage{cvpr}
\usepackage{times}
\usepackage{epsfig}
\usepackage{graphicx}
\usepackage{amsmath}
\usepackage{amssymb}
\usepackage{algorithm}
\usepackage{booktabs}
\usepackage{subfig}
\usepackage[noend]{algpseudocode}
\usepackage{multirow}
\usepackage{threeparttable}
\let\vec\boldsymbol

\usepackage{hyperref}

\cvprfinalcopy 

\def\cvprPaperID{6808} 

\ifcvprfinal\fi
\begin{document}

\title{NeuralScale: Efficient Scaling of Neurons for Resource-Constrained Deep Neural Networks}

\author{Eugene Lee \qquad Chen-Yi Lee\\
Institute of Electronics, National Chiao Tung University\\
{\tt\small eugenelet.ee06g@nctu.edu.tw \qquad cylee@si2lab.org}}

\maketitle
\thispagestyle{empty}

\begin{abstract}
   Deciding the amount of neurons during the design of a deep neural network to maximize performance is not intuitive. In this work, we attempt to search for the neuron (filter) configuration of a fixed network architecture that maximizes accuracy. Using iterative pruning methods as a proxy, we parameterize the change of the neuron (filter) number of each layer with respect to the change in parameters, allowing us to efficiently scale an architecture across arbitrary sizes. We also introduce architecture descent which iteratively refines the parameterized function used for model scaling. The combination of both proposed methods is coined as NeuralScale. To prove the efficiency of NeuralScale in terms of parameters, we show empirical simulations on VGG11, MobileNetV2 and ResNet18 using CIFAR10, CIFAR100 and TinyImageNet as benchmark datasets. Our results show an increase in accuracy of 3.04\%, 8.56\% and 3.41\% for VGG11, MobileNetV2 and ResNet18 on CIFAR10, CIFAR100 and TinyImageNet respectively under a parameter-constrained setting (output neurons (filters) of default configuration with scaling factor of 0.25).
\let\thefootnote\relax\footnotetext{Work supported by MOST of Taiwan: 107-2221-E-009-125-MY3.}
\end{abstract}

\section{Introduction}


\begin{figure}[!tbp]
\centering
  	\subfloat[NeuralScale.]{\includegraphics[width=0.3\textwidth]{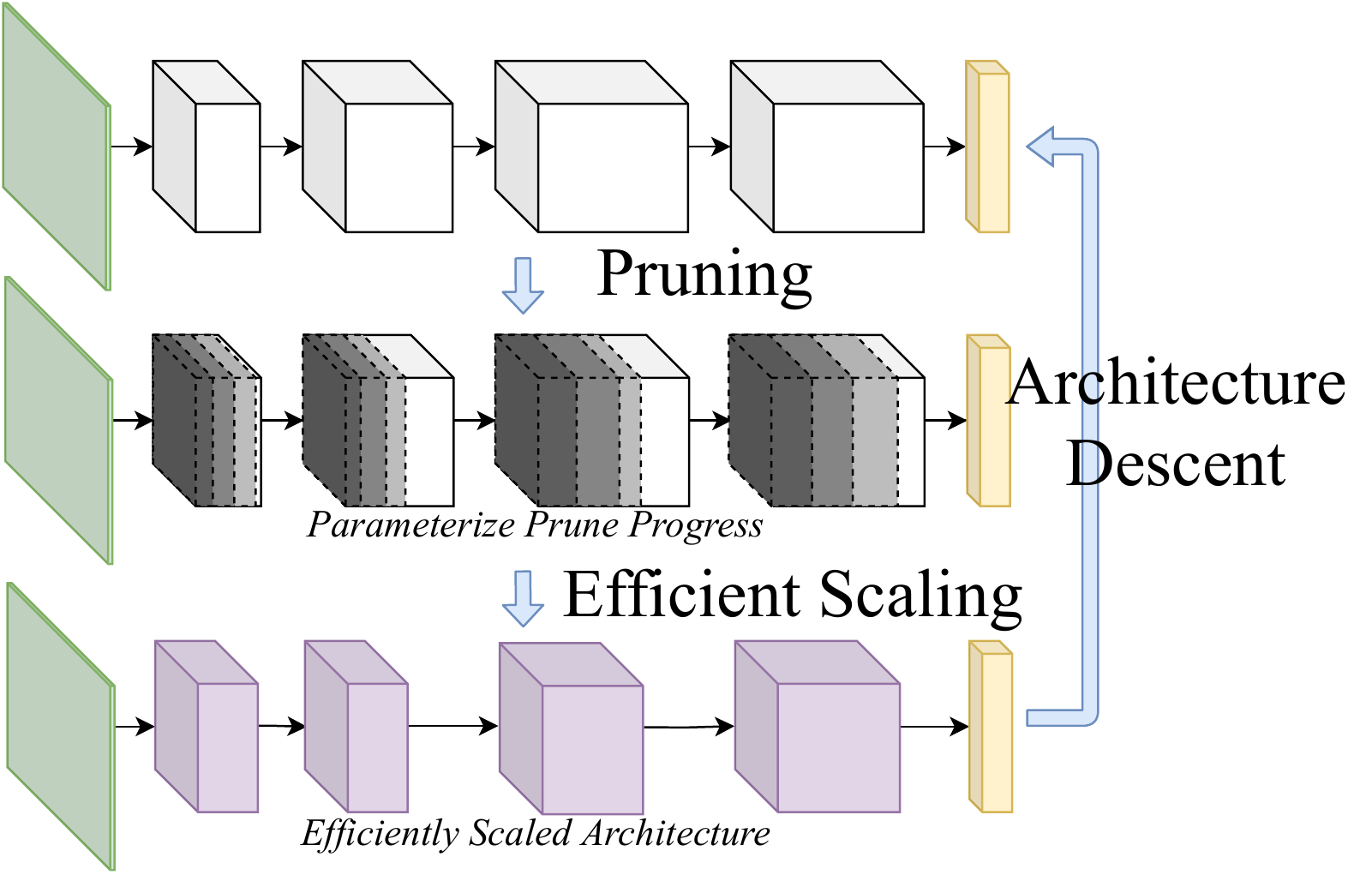}\label{fig:neuralscale}}
	\hfill
  	\subfloat[Uniform Scaling.]{\includegraphics[width=0.3\textwidth]{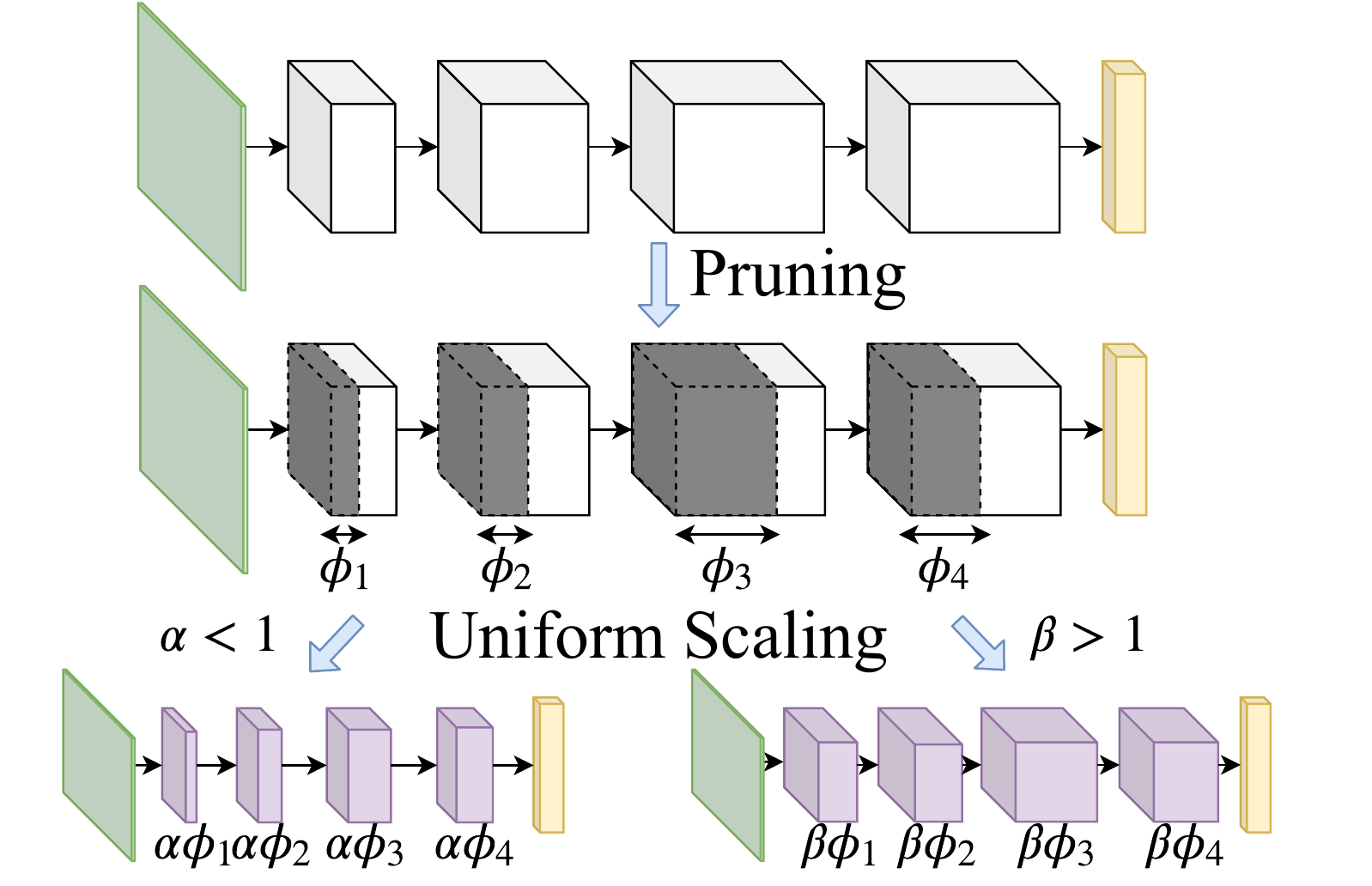}\label{fig:uniform_scale}}
\caption{
The input is represented as a green slab on the left, the output layer is the yellow bar on the right and intermediate layers are represented by 3D blocks with the width as its neuron (filter) number. The purple blocks are the final configuration of the neurons (filters). \protect\subref{fig:neuralscale} shows our proposed method that non-linearly scales neurons (filters) across different layers to maximize performance. \protect\subref{fig:uniform_scale} is a uniform scaling method that is shown to be effective in \cite{gordon2018morphnet}.}
\label{fig:scaling_methods}
\end{figure}

The human brain contains around 100 billion of neurons \cite{herculano2009human} that are structured in such a way that they are utilized in an efficient manner. As the design of deep neural network (DNN) is inspired by the human brain, there's one key ingredient that is missing from the current design of DNNs: the efficient utilization of resources (parameters).

The success of DNN is a composition of many factors. On an architectural level, various architectures have been proposed to increase the accuracy of DNNs targeting efficiency in computational cost (FLOPs) and size (parameters). In a modern DNN architectures, hyperparameters like width (neurons/filters), depth, skip-connections and activation functions \cite{nair2010rectified,ramachandran2017searching} are the building blocks. Notable architectures that are constructed using those building blocks are: VGGNet \cite{simonyan2014very}, ResNet \cite{he2016deep}, DenseNet \cite{huang2017densely}, GoogLeNet \cite{szegedy2015going} and MobileNets \cite{howard2017mobilenets,sandler2018mobilenetv2,howard2019searching}. Apart from the advances in architecture design, initialization of weights also helps in improving the accuracy of a DNN \cite{glorot2010understanding,he2015delving}.

We focus on optimizing the configuration of convolutional neural networks (CNNs) and shed light on the selection of the number of filters for each layer given a fixed architecture and depth. Our approach is complementary to the modern variants of CNNs (VGGNet, ResNet, MobileNet, etc.) through the introduction of a guided approach in tuning its width instead of just blindly stacking additional layers to boost accuracy. Our approach also investigates the conventional wisdom on filter selection stating that as we go deeper into the network, more filters are required to capture high-level information embedded in the features and to compensate with the gradual reduction in the spatial dimension which has efficiency in FLOPs as a byproduct.

Intuitively speaking, the design of traditional CNNs is ad-hoc and introduces redundancy \cite{li2016pruning,cogswell2015reducing}. This redundancy gives the opportunity for filter pruning techniques \cite{han2015learning,han2015deep,li2016pruning,liu2017learning} to strive, by conserving or improving accuracy using a lower parameter count. Current advances in pruning has led to a recent work by Liu \etal \cite{liu2018rethinking} that studies pruning in a new perspective. They show that pruning can be viewed as an architecture search method instead of just for removing redundancy. We incorporated this perspective along with the recent findings of EfficientNet \cite{tan2019efficientnet} stating that through the search of an optimal ratio between the width, depth and resolution of a given architecture and dataset, the accuracy of a network can improve if scaled accordingly. Both these works led us to think that instead of finding the scale or ratio among the width, depth and resolution, we can scale the width of a CNN across several layers independently using global iterative pruning as a proxy. We hypothesize that if we are given a DNN with minimal redundant parameters, through the modeling of the change of neurons (filters) in each layer of the DNN with respect to the change in the total parameters of the DNN, we are able to scale the DNN across various sizes efficiently. Our approach can also be viewed as a variant of neural architecture search (NAS) where the search is on finding the optimal configuration of neurons (filters) across layers instead of searching for the optimal structure involving skip-connections or filter types \cite{zoph2016neural,liu2018progressive,pham2018efficient,xie2018snas,bello2017neural,liu2018darts}. Our approach is comparatively light-weight as the only resource intensive task lies on the pruning of network. The gist of our proposed method is shown in Figure \ref{fig:neuralscale}.

The rest of the paper is structured as follows. We first show related work on available pruning techniques and the role of pruning for neural architecture search in Section \ref{sec:related_work}. We then show the details of our approach in Section \ref{sec:method}. Extensive experiments on our proposed method is shown in Section \ref{sec:experiments}. We finally conclude our paper in Section \ref{sec:conclusion}.
\section{Related Work}\label{sec:related_work}
\paragraph{Pruning of Deep Neural Network.}
Pruning of neural networks has been studied for decades with the goal of parameter reduction \cite{lecun1990optimal,hassibi1993second}. Parameter reduction is important if a DNN needs to be deployed to targeted devices with limited storage capacity and computational resources. There are two ways to prune a network: structured pruning \cite{liu2017learning,he2019filter,neklyudov2017structured,lee2018snip,luo2017thinet,zhao2019variational} and unstructured pruning \cite{han2015deep,han2016dsd,han2015learning,frankle2018lottery}. For structured pruning, entire filters or neurons are removed from a layer of a network. Such pruning method can be deployed directly to a general computing architecture, e.g.\ GPU, to see improvement in speed. For unstructured pruning, individual weights are pruned, resulting in a sparse architecture. A dedicated hardware is required to exploit the speed-up contributed by the sparsity of weights.

To prune a network, there are various criteria that have been studied. The most intuitive approach is to prune the weights based on its magnitude \cite{han2015learning,han2015deep,li2016pruning}. It was believed that the importance of weights is related to its magnitude. Although this approach is widely used in other works \cite{frankle2018lottery,li2016pruning,he2018amc,yang2017designing}, it is also shown on several works \cite{molchanov2019importance,he2019filter,lecun1990optimal,hassibi1993second} that magnitude pruning does not result in an architecture with the best performance. Pruning based on magnitude is adopted because of its simplicity when compared to more complicated pruning techniques, e.g.\ \cite{hassibi1993second,lecun1990optimal} requires the computation of Hessian for pruning. A study proposed the use of geometric median \cite{he2019filter} as a replacement of magnitude pruning for the criteria of network pruning. \cite{molchanov2019importance,molchanov2016pruning} has also challenged the reliability of magnitude pruning and proposed the use of Taylor expansions to approximate the contribution of a filter to the final loss.  Another intuitive way of pruning is through the addition of a regularizer to the loss function to force redundant weights to zero during training \cite{han2015learning}. It has also been discovered that the scaling parameter used in Batch Normalization (BN) \cite{ioffe2015batch} can be used for structured pruning and yields performance better than pruning using magnitude \cite{liu2017learning}. A follow-up work that takes the shift term of BN into consideration for pruning is proposed in \cite{zhao2019variational}.

\paragraph{Neural Architecture Search via Pruning.}
There's a tremendous surge of efforts placed into the research of neural architecture search (NAS) techniques in the recent years on coming up with the most efficient architecture possible for a given task \cite{zoph2016neural,liu2018progressive,pham2018efficient,tan2019mnasnet,cai2018proxylessnas,liu2018darts}. NAS techniques are usually computationally expensive, limiting its applicability to research or corporate environment with limited computing resources. The search space of NAS is very broad and is defined distinctively across different works. Most of the search space involves the search of a suitable set of operations to be placed in a cell. The connections between different operations is also considered in the search space \cite{bender2018understanding}. These cells are then stacked to increase network depth.

In our work, we focus on the decision of the number of neurons (filters) required for each layer in a DNN. We use existing pruning techniques as a proxy to tackle this problem. The idea of using pruning as an architecture search method is not novel and has been discussed in \cite{liu2018rethinking,frankle2018lottery,frankle2019lottery,ding2019approximated} where its applicability can be seen in MorphNet \cite{gordon2018morphnet}. Liu \etal \cite{liu2018rethinking} show that through pruning, we are removing redundancy from a network and the resulting network is efficient in terms of parameters. They also show that training the pruned architecture from scratch has comparable, if not better, accuracy than fine-tuned networks, indicating that the accuracy gain is from the resulting efficient architecture obtained via pruning. For the case of unstructured pruning, it is studied in \cite{frankle2018lottery,frankle2019lottery,zhou2019deconstructing} that a Lottery Ticket (LT) can be found via iterative unstructured pruning. A LT is a sparse architecture that is the result of unstructured pruning and has accuracy better than the original network (usually found at a parameter count of an order less than the parameter count of the original network). This finding indicates that pruning does introduce inductive bias \cite{cohen2016inductive} and adds evidence on the suitability of using pruning as an architecture search technique. This idea is proven in MorphNet \cite{gordon2018morphnet} where a pruned architecture is scaled uniformly to meet the targeted parameter count and is repeated for several iterations. A single iteration of \cite{gordon2018morphnet} is illustrated in Figure \ref{fig:uniform_scale}.

\section{Method}\label{sec:method}
\subsection{Parameter Tracking via Iterative Pruning}
To efficiently allocate neurons (filters) across different layers of a DNN that results in optimal accuracy given a parameter constraint, we model the change of neurons (filters) across layers with respect to the change of parameters. First, we need to begin with a network with minimal redundant parameters. To do so, a structured pruning method proposed by Molchanov \etal \cite{molchanov2019importance} that prunes iteratively is adopted. They proposed a pruning method that prunes neurons based on its importance. The importance of a parameter can be measured as the loss induced when it's removed from the network. They proposed the use of Taylor approximation as an efficient way to find parameters that are of less importance. A comprehensive comparison between their approach and an oracle (full combinatorial search over all parameters that results in minimum increase in loss) is done, proving its reliability. Here, we will give a brief introduction of their parameter pruning technique borrowed from their paper. The importance of a parameter is quantified by the error induced when it is removed from the network:
\begin{equation}
    \mathcal{I}_m = (E(\mathcal{D},\vec{W}) - E(\mathcal{D},\vec{W}|w_m = 0))^2. \label{eq:importance}
\end{equation}
Here, $\vec{W}=\{w_0,w_1,...,w_M\}$ are the set of parameters of a neural network supported by a dataset $\mathcal{D}=\{(x_0,y_0),(x_1,y_1),...,(x_K,y_K)\}$ of $K$ independent samples composed of inputs $x_k$ and outputs $y_k$. (\ref{eq:importance}) can be approximated by the second-order Taylor expansion as:
\begin{equation}
    \mathcal{I}_m^{(2)}(\vec{W}) = (g_m w_m - \frac{1}{2} w_m \vec{H}_m \vec{W})^2. \label{eq:second_order}
\end{equation}
$\vec{H}$ is the Hessian matrix where $\vec{H}_m$ is the $m$-th row of it and $g_m=\frac{\partial E}{\partial w_m}$. (\ref{eq:second_order}) can be further approximated using the first-order expansion:
\begin{equation}
    \mathcal{I}_m^{(1)}(\vec{W}) = (g_m w_m)^2. \label{eq:first_order}
\end{equation}
To minimize computational cost, (\ref{eq:first_order}) will be used since it is shown in \cite{molchanov2019importance} that the performance is on par with the second-order expansion and the first-order Taylor expansion is often used to estimate the importance of DNN components (weights, kernels or filters) \cite{theis2018faster,molchanov2016pruninginfer,ding2019global}. Consistent with their work, a gate $\vec{z}_m$ is placed after batch normalization layers \cite{ioffe2015batch} where the importance approximation is defined as:
\begin{equation}
    \mathcal{I}_m^{(1)}(\vec{z}) = (\frac{\partial E}{\partial \vec{z}_m})^2 = (\sum_{s\in \mathcal{S}_m} g_m w_m)^2, \label{eq:criteria}
\end{equation}
where $\mathcal{S}_m$ corresponds to the set of weights $\vec{W}_{s\in \mathcal{S}_m}$ before the gate.

For a network composed of $L$ layers, we define the the set of neurons (filters) for the entire network as $\{\phi_l\}_{l=1}^L$. $\phi_l$ is the number of neurons (filters) of layer $l$. We then define the total number of parameters in a network as $\tau$. As we are using an iterative pruning method, on every pruning iteration, we will obtain a set of $\tau$'s and $\phi_l$'s for the $l$-th layer which can be represented as $\xi_l = \{\tau,\phi_l\}$. After pruning for $N$ iterations, we obtain $\vec{\xi}_l = \{\{\tau^{(n)},\phi_l^{(n)}\}_{n=1}^N\}$. In our implementation, we start feeding $\xi_l$'s into $\vec{\xi}_l$ when all layers in a network is pruned by at least a single parameter. We conjecture that when all layers are pruned by at least a single parameter, most redundancy is removed and the residual parameters compose an efficient configuration. We stop pruning once the number of neurons (filters) is less than $\epsilon$ (we pick $\epsilon$ as 5\% of the total neurons (filters) of the network in our implementation). Upon the completion of pruning, we have $\vec{\xi}\in \mathbb{R}^{2\times N\times L}$. The entire pruning process begins once we have trained our network for $P$ epochs (commonly known as network pre-training; we use the term \textit{pre-training epochs} in our context instead) using a learning rate $\mu$. The choice of $P$ is studied and the conventional wisdom on the requirement of pre-training a network to convergence before pruning is investigated in the Supp.\ Section \ref{sec:pre-training}. The pruning algorithm is summarized in Algorithm \ref{alg:iterative_prune}.

\begin{algorithm}
\caption{Iterative Prune}\label{alg:iterative_prune}
\begin{algorithmic}[1]
\Procedure{IterativePrune}{$f,\mathcal{D}$}
\For{$P$ epochs}
    \State Update $f$ using learning rate $\mu$
\EndFor
\While{$\sum_{l=1}^L |\vec{z}_l|_1 >\epsilon$}
    \State $\text{Train $f$ for $Q$ iterations}$
    \State $\vec{\xi}' \gets \text{Prune $f$ using criteria (\ref{eq:criteria})}$
    \If{\text{all layers pruned at least once}}
        \State $\vec{\xi} \gets \{\vec{\xi},\vec{\xi}' \}$ \Comment{Record parameters}
    \EndIf
\EndWhile

\State \textbf{return} $\vec{\xi}$
\EndProcedure
\end{algorithmic}
\end{algorithm}

\subsection{Efficient Scaling of Parameters}\label{sec:efficient_scaling_of_parameters}

The goal of this work is to scale the neurons (filters) of a network across different layers to satisfy the targeted total parameter size denoted as $\hat{\tau}$. For parameter scaling to match the targeted size, uniform scaling is used in MorphNet \cite{gordon2018morphnet} and MobileNets \cite{howard2017mobilenets} where all layers are scaled with a constant width multiplier. It is intuitive that the scale applied to neurons (filters) of different layers should be layer-dependent to maximize performance. In this work, we propose an efficient method to scale the number of neurons (filters) across different layers to maximize performance. By using iterative pruning as a proxy, we parameterize the change of neurons (filters) across different layers with respect to the total parameters or simply put as $\frac{\partial \phi_l}{\partial\tau}$.

\paragraph{Modeling Parameter Growth.}

\begin{figure}[!tbp]
\centering
  	\subfloat[CIFAR10.]{\includegraphics[width=.5\columnwidth]{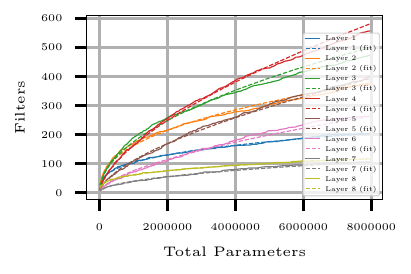}\label{fig:growth_c10}}
	\hfill
	\subfloat[CIFAR100.]{\includegraphics[width=.5\columnwidth]{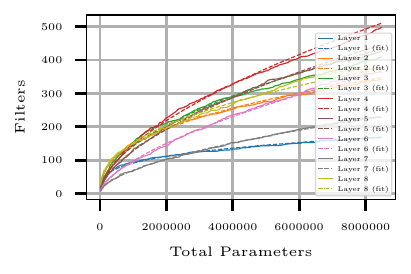}\label{fig:growth_c100}}
\caption{Growth in number of filters of different layers across various network sizes. Each color represents independent layers of the convolutional filters of VGG11. Solid line is the residual filters obtained using an iterative pruning method and dashed line represents our approach on curve fitting. It can be observed that there's a pattern in the change of filters w.r.t.\ parameters which is dataset dependent.}
\label{fig:param_growth}
\end{figure}

As discussed in the previous subsection, $\vec{\xi}_l$ collected for each layer resembles the efficient set of neurons (filters) for each layer at a given size constraint. We can use this as a proxy to model $\frac{\partial \phi_l}{\partial \tau}$. We first observe how the distribution of the residual neurons (filters) obtained using pruning scale across different $\tau$'s, e.g.\ we use VGG11 as our network and CIFAR10/100 as our dataset to show the parameter growth across various sizes in Figure \ref{fig:param_growth}. We can observe that parameters across different layers don't scale linearly across different sizes, implying that uniform scaling is not an efficient scaling method. Figure \ref{fig:param_growth} also shows us that the growth of the parameters resembles a power function that is monotonic. To fit the curves, we use:
\begin{equation}
    \phi_l (\tau | \alpha_l , \beta_l) = \alpha_l \tau ^{\beta_l}, \label{eq:phi_l}
\end{equation}
where every layer is paramterized independently by $\alpha_l$ and $\beta_l$. To obtain these parameters, we can linearize the problem by taking $\ln$ on both sides of (\ref{eq:phi_l}), giving us:
\begin{equation}
    \ln \phi_l (\tau | \alpha_l, \beta_l) = \ln \alpha_l + \beta_l \ln \tau.
\end{equation}
Since we pruned the network iteratively for $N$ iterations, we obtain a set of linear equations which can  be formulated into a matrix of the form:
\begin{align} \label{eq:matrix}
\underbrace{
\begin{bmatrix} 
    1 & \ln\tau^{(1)} \\
    1 & \ln\tau^{(2)} \\
    \vdots  & \vdots \\
    1 & \ln\tau^{(N)} 
\end{bmatrix}}_{\vec{\mathcal{T}}}
\underbrace{
\begin{bmatrix} 
    \ln\alpha_1 & \ln\alpha_2 & \dots & \ln\alpha_L\\
    \beta_1 & \beta_2 & \dots & \beta_L
\end{bmatrix}}_{\vec{\Theta}} = \\  \nonumber
\hfill \underbrace{\begin{bmatrix} 
    \ln\phi_1^{(1)} & \ln\phi_2^{(1)} & \dots & \ln\phi_L^{(1)} \\
    \ln\phi_1^{(2)} & \ddots & &\\
    \vdots & & & \\
    \ln\phi_1^{(N)} &   &     & \ln\phi_L^{(N)}   
\end{bmatrix}}_{\vec{\Phi}}.
\end{align}
We can solve for $\vec{\Theta}$ in (\ref{eq:matrix}) using the least-squares approach or by taking the pseudoinverse of $\vec{\mathcal{T}}$ and multiply it with $\vec{\Phi}$:
\begin{equation}
    \vec{\Theta} = (\vec{\mathcal{T}}^T \vec{\mathcal{T}})^{-1}\vec{\mathcal{T}}^T\vec{\Phi}. \label{eq:pseudo}
\end{equation}
By using this method, we are able to fit the curves or $\vec{\xi}_l$'s obtained using iterative pruning of a network where the fitted results are shown in Figure \ref{fig:param_growth}. Our approach is a cost effective way of neural architecture parametrization and only takes two parameters ($\vec{\alpha}$ and $\vec{\beta}$) per layer to parameterize the non-linear growth of the neuron (filter) count across various parameter sizes or $\frac{\partial \phi_l}{\partial \tau}$. The simplicity of our approach also prevents the overfitting of noise embedded in the samples obtained via pruning. The search of parameters for efficient scaling is summarized in Algorithm \ref{alg:search_parameters}.

\begin{algorithm}
\caption{Search Parameters}\label{alg:search_parameters}
\begin{algorithmic}[1]
\Procedure{SearchParams}{$\vec{\xi}$}
\State $\mathcal{T},\vec{\Phi} \gets \vec{\xi}$ \Comment{Convert to matrix form}
\State $\vec{\Theta} \gets (\vec{\mathcal{T}}^T \vec{\mathcal{T}})^{-1}\vec{\mathcal{T}}^T\vec{\Phi} $
\State \textbf{return} $\vec{\Theta}$
\EndProcedure
\end{algorithmic}
\end{algorithm}

\paragraph{Meeting Parameter Constraints.}
Since our approach fully parameterizes the independent scaling of network width across various sizes, we can meet tight parameter constraints during deployment of a DNN to devices with limited resource budget. For approaches like uniform scaling \cite{gordon2018morphnet,liu2018rethinking,howard2017mobilenets}, only the number of output neurons (filters) can be scaled while the network size is a function of the input and output neurons (filters), hence meeting parameter constraints can only be done by performing an iterative grid search on the number of output neurons (filters) required. 

For our approach, parameter scaling is intuitive as we can apply gradient descent on $\vec{\Phi}$ w.r.t.\ $\tau$. To do so, we define a DNN as $f(\vec{x}|\vec{W},\vec{\Phi(\tau|\vec{\Theta})})$ where $f$ is a DNN architecture, $\vec{x}$ is its input, $\vec{W}$ are the weights of the DNN, and the additional condition $\vec{\Phi(\tau|\vec{\Theta})}$ is introduced to parameterize the number of neurons (filters) required for each layer of a DNN. We then define a function $h$ that computes the number of parameters of a DNN. $h$ is architecture dependent. 

Given a parameter constraint $\hat{\tau}$ that needs to be met, we can generate an architecture having total parameters close to $\hat{\tau}$ by performing stochastic gradient descent (SGD) on $\vec{\Phi}(\tau,\vec{\Theta})$ w.r.t.\ $\tau$. Like other gradient descent problems, parameter initialization is important and the best way to do so is to fit $\hat{\tau}$ into (\ref{eq:phi_l}) giving us:
\begin{equation}
    \phi_l(\hat{\tau}| \alpha_l, \beta_l) = \alpha_l \hat{\tau} ^{\beta_l}.
\end{equation}
This gives us a good initial point, however there will still be a discrepancy between $h(f(\vec{x}|\vec{W},\vec{\Phi(\hat{\tau}|\vec{\Theta})}))$ and  $\hat{\tau}$ which can be fixed by applying SGD on $\frac{1}{2}(\phi_l(\tau| \alpha_l, \beta_l) - \hat{\tau})^2$ w.r.t. to $\tau$ where the update of $\tau$ is given as:
\begin{align}
    \tau_i &= \tau_{i-1} - \Delta \tau_{i-1} \\
        &= \tau_{i-1} - \eta(h(f(\vec{x}|\vec{W},\vec{\Phi}(\tau_{i-1}|\vec{\Theta}))) - \hat{\tau}  )\sum_{l=1}^L \beta_l \alpha_l \tau_{i-1}^{\beta_l - 1}. \label{eq:tau_update}
\end{align}
Here, the subscript of $\tau$ corresponds to the SGD iteration and the full proof of (\ref{eq:tau_update}) is given in Supp.\ Section \ref{sec:update_tau}. Since the number of parameters of an architecture is a monotonic function of $\Phi$, this problem is convex and converges easily. If the learning rate $\eta$ is set carefully, we are able to obtain an architecture with total parameters close to $\hat{\tau}$. We summarize this procedure in Algorithm \ref{alg:generate_network}.

\begin{algorithm}
\caption{Generate Network Using Searched Parameters}\label{alg:generate_network}
\begin{algorithmic}[1]
\Procedure{GenerateNet}{$\vec{\Theta},f,\hat{\tau}$}
\State $\tau \gets \hat{\tau}$ \Comment{Initialize parameter}
\While{not converged}
\State $\tau\gets \tau - \Delta \tau$\Comment{Update using SGD as in (\ref{eq:tau_update})}
\EndWhile
\For{$l \gets 1$ to $L$} \Comment{Layer-wise architecture update}
\State $f_l\gets $ set output neurons (filters) as $\phi_l(\tau|\vec{\Theta}_l)$ (\ref{eq:phi_l})
\EndFor
\State \textbf{return} $f$
\EndProcedure
\end{algorithmic}
\end{algorithm}

\subsection{Architecture Descent for Model Refinement}
Like any gradient descent algorithm, initialization plays an important role and affects the convergence of an algorithm. Our approach is similar in way where we attempt to search for the configuration of an architecture given an initial configuration, e.g.\ VGGNet \cite{simonyan2014very} and ResNet \cite{he2016deep} consist of a set of predefined filter numbers for different configurations. As our approach behaves similarly to gradient descent, we coin it \textit{architecture descent} as there is no gradient involved and it is descending in the loss surface by making iterative changes to the architecture's configuration.

We define an iteration of architecture descent as a single run of Algorithm \ref{alg:iterative_prune}, \ref{alg:search_parameters} and \ref{alg:generate_network} that corresponds to iterative pruning, parameter searching and network generation. Upon the completion of iterative pruning and parameter searching, we obtain a set of parameters that scales our network in a more efficient manner. We can then use this set of parameters to scale-up our network as shown in Algorithm \ref{alg:generate_network} for further pruning. We then proceed with several iterations of architecture descent until the changes in the architecture configuration is minuscule, indicating convergence. By applying architecture descent, we are descending on the loss surface that is parameterized by $\vec{\Theta}$ instead of a loss surface parameterized by its weights $\vec{W}$ performed in gradient descent. Architecture descent is summarized in Algorithm \ref{alg:architecture_descent}. NeuralScale is a composition of all algorithms we proposed as illustrated in Figure \ref{fig:neuralscale}.

\begin{algorithm}
\caption{Architecture Descent}\label{alg:architecture_descent}
\begin{algorithmic}[1]
\Procedure{ArchitectureDescent}{$f,\mathcal{D},\tau$}
\While{not converged}
\State $\vec{\xi}\gets \textsc{IterativePrune(}f,\mathcal{D}\textsc{)}$\Comment{Taylor FO \cite{molchanov2019importance}}
\State $\vec{\Theta}\gets \textsc{SearchParams(}\vec{\xi}\textsc{)}$
\State $f\gets \textsc{GenerateNet(}\vec{\Theta},f,\tau\textsc{)}$
\EndWhile
\State \textbf{return} $f$\Comment{Network with scaled parameters}
\EndProcedure
\end{algorithmic}
\end{algorithm}

\section{Experiments}\label{sec:experiments}
In this section, we show experiments illustrating the importance of architecture descent. We then proceed with the benchmarking of our approach using public datasets, e.g.\ CIFAR10/100 \cite{krizhevsky2009learning} and TinyImageNet (subset of ImageNet \cite{deng2009imagenet} with images downsampled to $64\times64$ and consists of 200 classes having 500 training and 50 validation samples for each class) on commonly used architectures, e.g.\ VGG11 \cite{simonyan2014very}, MobileNetV2 \cite{sandler2018mobilenetv2} and ResNet18 \cite{he2016deep}. All experiments are run on a single GTX1080Ti GPU using PyTorch \cite{paszke2017automatic} as our deep learning framework. We use SGD as our optimizer with an initial learning rate of 0.1, momentum set to 0.5 and a weight decay factor of $5^{-4}$. Training of network that uses CIFAR10 and CIFAR100 are run for 300 epochs using a step decay of learning rate by at factor of 10 at epochs 100, 200 and 250 whereas network trained using TinyImageNet are run for 150 epochs with a decay in learning rate by a factor of 10 at epochs 50 and 100. For iterative pruning, we first train our network for $P=10$ epochs using a learning rate of 0.1 and is decayed by a factor of 10 every 10 epochs.  Source code is made available at \url{https://github.com/eugenelet/NeuralScale}.

\subsection{Importance of Architecture Descent}\label{sec:importance_architecture_descent}
 For all experiments, we run architecture descent for 15 iterations. We show configurations with total parameters matching total parameters of network with its default set of filters uniformly scaled to a ratio, $r$. $r=0.25,2$ for CIFAR10 and CIFAR100. $r=0.25,1$ for TinyImageNet.
 
\begin{figure}[!tbp]
\centering
\includegraphics[width=.9\columnwidth]{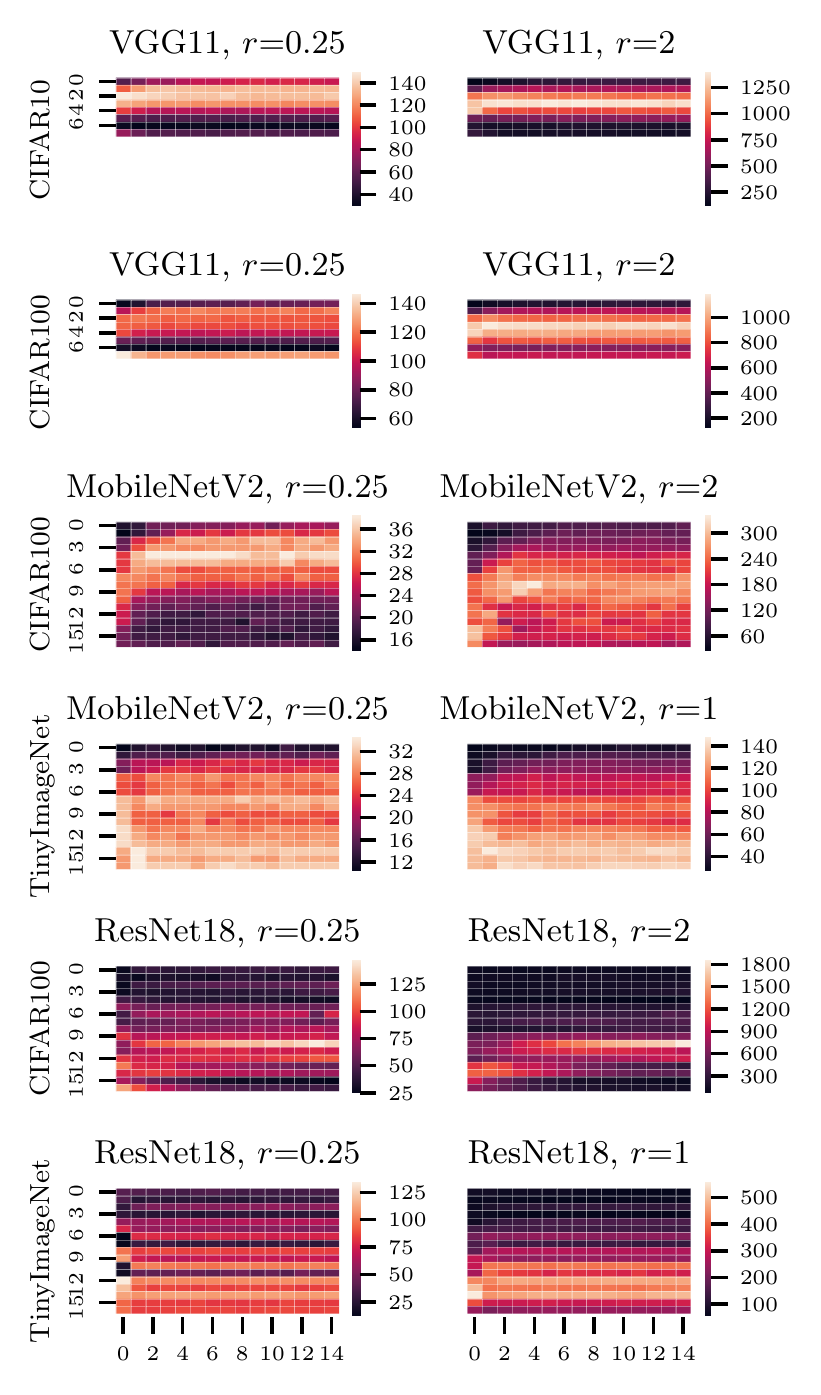}\label{fig:vgg11_c10_arch}
\caption{Shows the number of filters for each layer by running architecture descent for 15 iterations on various architecture-dataset pair. Vertical and horizontal axis of each plot corresponds to the filter number and architecture descent iteration respectively. $r$ is the uniform scaling ratio applied to the default configuration. Best viewed in color.}
\label{fig:architecture_descent}
\end{figure}

 \paragraph{VGG11.} Using a relatively shallow network, we demonstrate the application of architecture descent using CIFAR10 and CIFAR100 as shown in Figure \ref{fig:architecture_descent}. By observing the resulting architecture configuration, we can make two conjectures. First, we show that conventional wisdom on network design that gradually increases the number of filters as we go deeper in a convolutional network does not guarantee optimal performance. It can be observed that the conventional wisdom on network design holds up to some level (layer 4) and bottlenecking of parameters can be observed up to the penultimate layer which is followed by a final layer which is comparatively larger. Second, the scaling of network should not be done linearly as was done in \cite{howard2017mobilenets} and should follow a non-linear rule that we attempt to approximate using a power function. If we look closely, by applying architecture descent on datasets of higher complexity generates network with configuration that has more filters allocated toward the end. Our conjecture is that more resources are needed to capture the higher level features when the task is more difficult whereas for simple classification problem like CIFAR10, more resource is allocated to earlier layers to generate more useful features. These observations give us a better understanding on how resource should be allocated in DNNs and can be used as a guideline for deep learning practitioners in designing DNNs. A single iteration of architecture descent for VGG11 on CIFAR10/100 is approximately 20 minutes.
 
\paragraph{MobileNetV2.} We show the application of architecture descent on a more sophisticated architecture known as MobileNetV2 using CIFAR100 and TinyImageNet in Figure \ref{fig:architecture_descent}. Here, we only apply our search algorithm on deciding the size of the bottleneck layers while the size of the expansion layer follows the same expansion rule found in \cite{sandler2018mobilenetv2} where an expansion factor of $6\times$ is used. The resulting configuration closely resembles the one found using a feedforward network like VGG. It can also be observed that resources are allocated toward the output for a more sophisticated dataset. A single iteration of architecture descent for MobileNetV2 on CIFAR100 and TinyImageNet is approximately 50 minutes and 1.2 hour respectively.

 \paragraph{ResNet18.} The application of architecture descent on ResNet18 using CIFAR100 and TinyImageNet is shown in Figure \ref{fig:architecture_descent}. We observe a different pattern of architecture configuration when compared to a simple feed forward network like VGG. This is an interesting observation as it agrees with the interpretation of residual networks as an ensemble of many paths of different lengths shown in \cite{veit2016residual}. Another observation is that if we look only at a single layer of every residual block (each block consists of two layers), the searched configuration for ResNet follows the pattern found in VGG where there's a smooth gradient of filter progression across layers. A single iteration of architecture descent for ResNet18 on CIFAR100 and TinyImageNet is approximately 50 minutes and 45 minutes respectively.

\subsection{Benchmarking of NeuralScale}
\begin{figure}[!tbp]
\centering
  	\subfloat[VGG11 on CIFAR10.]{\includegraphics[width=0.235\textwidth]{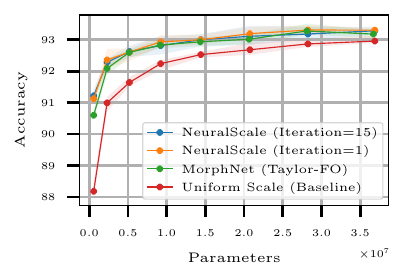}\label{fig:vgg11_c10}}
	\hfill
	\subfloat[VGG11 on CIFAR100.]{\includegraphics[width=0.235\textwidth]{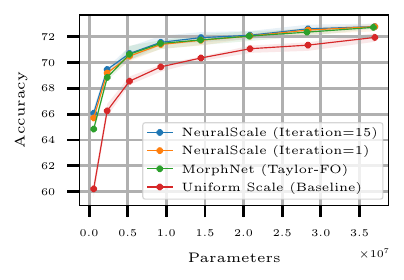}\label{fig:vgg11_c100}}
	\vfill
  	\subfloat[MobileNetV2 on CIFAR100.]{\includegraphics[width=0.235\textwidth]{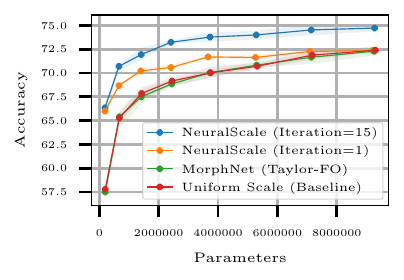}\label{fig:mobilev2_c100}}
	\hfill
	\subfloat[MobileNetV2 on TinyImageNet.]{\includegraphics[width=0.235\textwidth]{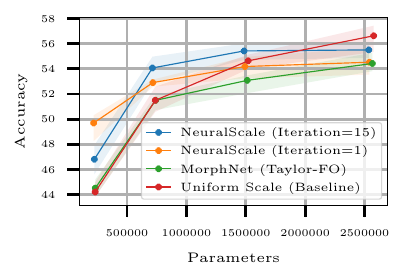}\label{fig:mobilev2_imagenet}}
	\vfill
  	\subfloat[ResNet18 on CIFAR100.]{\includegraphics[width=0.235\textwidth]{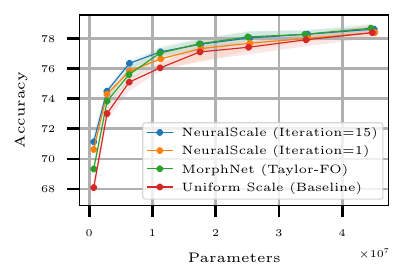}\label{fig:res18_c100}}
	\hfill
	\subfloat[ResNet18 on TinyImageNet.]{\includegraphics[width=0.235\textwidth]{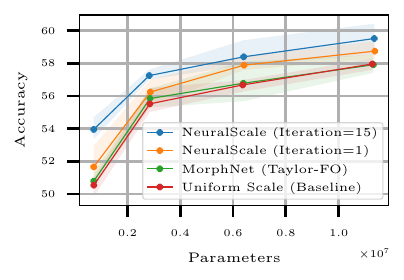}\label{fig:res18_imagenet}}
\caption{All plots are organized such that each row corresponds to a single architecture, e.g.\ \protect\subref{fig:vgg11_c10},\protect\subref{fig:vgg11_c100} corresponds to VGG11, \protect\subref{fig:mobilev2_c100},\protect\subref{fig:mobilev2_imagenet} corresponds to MobileNetV2 and \protect\subref{fig:res18_c100},\protect\subref{fig:res18_imagenet} corresponds to ResNet18. Each plot consists of the accuracy comparison of different scaling methods (applied on the width), plotted against different parameter scales. The shaded region of each line corresponds to the maximum and minimum accuracy across 5 independent simulations. }
\label{fig:accuracy_comparison}
\end{figure}

\begin{figure}[!tbp]
\centering
  	\subfloat[VGG11 on CIFAR10.]{\includegraphics[width=0.235\textwidth]{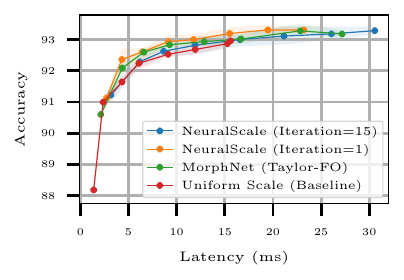}\label{fig:latency_vgg11_c10}}
	\hfill
	\subfloat[VGG11 on CIFAR100.]{\includegraphics[width=0.235\textwidth]{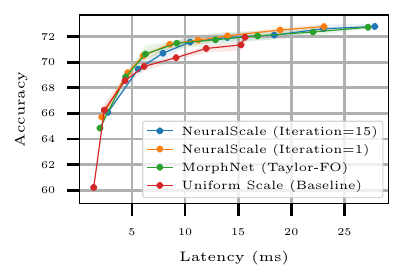}\label{fig:latency_vgg11_c100}}
	\vfill
  	\subfloat[MobileNetV2 on CIFAR100.]{\includegraphics[width=0.235\textwidth]{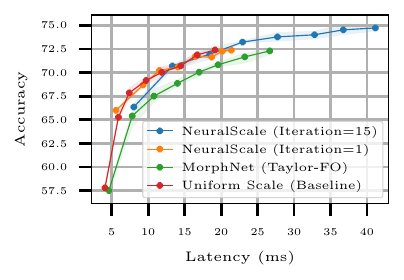}\label{fig:latency_mobilev2_c100}}
	\hfill
	\subfloat[MobileNetV2 on TinyImageNet.]{\includegraphics[width=0.235\textwidth]{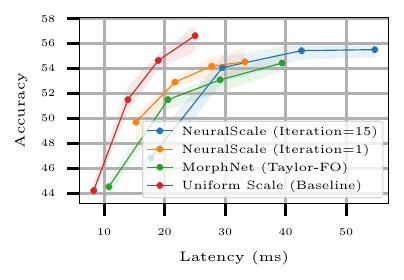}\label{fig:latency_mobilev2_imagenet}}
	\vfill
  	\subfloat[ResNet18 on CIFAR100.]{\includegraphics[width=0.235\textwidth]{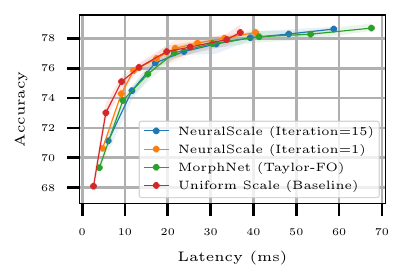}\label{fig:latency_res18_c100}}
	\hfill
	\subfloat[ResNet18 on TinyImageNet.]{\includegraphics[width=0.235\textwidth]{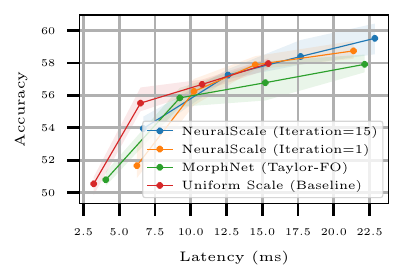}\label{fig:latency_res18_imagenet}}
\caption{The layout of these plots are structured as shown in Figure \ref{fig:accuracy_comparison}. The only difference is that the accuracies of different methods are plotted against latencies. }
\label{fig:latency_comparison}
\end{figure}

Here, we compare the accuracy of NeuralScale with the first (Iteration=1, to compare with MorphNet \cite{gordon2018morphnet}) and last (Iteration=15) iteration of architecture descent with a uniform scaling (baseline) method and a method where a network is first pruned using Taylor-FO \cite{molchanov2019importance} until it has 50\% of filters left and then scaled uniformly (resembling the first iteration of MorphNet \cite{gordon2018morphnet} and the use of \cite{molchanov2019importance} is to match our pruning method for a fair comparison), named as \textit{MorphNet (Taylor-FO)} in our comparison tables and plots. The accuracy is obtained by averaging across the maximum test accuracy of 5 independent simulations. For the accuracy plots in Figure \ref{fig:accuracy_comparison} and \ref{fig:latency_comparison}, the output filters of the original network are scaled to the ratios from 0.25 to 2 with an increment of 0.25 for CIFAR10/100 along with 0.25, 0.5, 0.75 and 1 for TinyImageNet. For the comparison tables, only the ratios 0.25, 0.75, 2 for CIFAR10/100 and 0.25, 0.75 for TinyImageNet are reported. Comparisons are also made with a structured magnitude pruning method \cite{li2016pruning} where we first pre-train our network using the same prescription for other methods and proceed with 40 and 20 epochs of fine-tuning for CIFAR10/100 and TinyImageNet respectively, using a learning rate of 0.001. We only show results for VGG11-CIFAR10, MobileNetV2-CIFAR100 and ResNet18-TinyImageNet in Table \ref{tab:acc_comparison} and the rest are deferred to Supp.\ Section \ref{sec:acc_comparison}. Note that all methods are trained from scratch and only \cite{li2016pruning} is trained using the pretrain-prune-finetuning pipeline. Results show that the hypothesis in \cite{liu2018rethinking} holds (training from scratch performs better). As our approach is designed for platforms with structured parallel computing capability like GPUs, we report the latency of different methods instead of FLOPs. Note that our approach isn't optimized for latency. Here, latency is defined as the time required for an input to propagate to the output. All latencies reported in Table \ref{tab:acc_comparison} are based on a batch size of 100 where 20 batches are first fed for warm-up of cache and is proceeded with 80 batches which are averaged to give the final latencies. As a comparison of latency based solely on the scale of parameter in Table \ref{tab:acc_comparison} is unintuitive, we show a plot comparing accuracy of different methods against latency in Figure \ref{fig:latency_comparison}.

\paragraph{VGG11.}
By observing the comparison plot shown in Figure \ref{fig:vgg11_c10} and \ref{fig:vgg11_c100}, our approach compares favourably in terms of parameter efficiency for CIFAR10 and CIFAR100. As shown in Table \ref{tab:acc_comparison}, at the lowest parameter count, an accuracy gain of 3.04\% is obtained for CIFAR10. Efficiency in latency of our approach is also comparable with the baseline approach as shown in Figure \ref{fig:latency_vgg11_c10} and \ref{fig:latency_vgg11_c100}. Diminishing returns are observed when the network increases in size. We conjecture that as the network grows larger, more subspaces are covered, hence the network can still adapt to the sub-optimal configuration by suppressing redundant parameters. Another observation is that the performance gain is more substantial on a more complicated dataset which is intuitive as inductive bias is introduced in an architectural level.

\begin{table}[t]
\begin{threeparttable}
  \caption{Comparison of various network-dataset pairs.}
  \label{tab:acc_comparison}
  \centering
  \begin{tabular}{cccc}
    \toprule
    Method & Params & Latency & Accuracy (\%) \\
    \midrule
    \midrule
    \multicolumn{4}{c}{\textbf{VGG11 CIFAR10}} \\
    \multirow{3}{*}{\parbox{2.3cm}{\centering Uniform Scale (Baseline)}} 
    & 0.58M & 1.29ms & 88.18 $\pm$ 0.16     \\
    & 5.20M & 4.31ms  & 91.64 $\pm$ 0.10 \\
    & 36.89M & 18.86ms  & 92.96 $\pm$ 0.09 \\
    \midrule
    Li \etal \cite{li2016pruning}\tnote{$\dagger$} 
    & 5.20M & 4.75ms  & 91.12 $\pm$ 0.02 \\
    \midrule
    \multirow{3}{*}{\parbox{2.4cm}{\centering MorphNet \cite{gordon2018morphnet} (Taylor-FO \cite{molchanov2019importance})}}            
    & 0.58M & 2.07ms & 90.60 $\pm$ 0.11     \\
    & 5.22M & 8.43ms  & 92.60 $\pm$ 0.09 \\                         
    & 36.72M & 48.52ms  & 93.18 $\pm$ 0.11 \\                                    
    \midrule
    \multirow{3}{*}{\parbox{2.3cm}{\centering NeuralScale (Iteration = 1)}}     
    & 0.58M & 2.56ms & 91.13 $\pm$ 0.07     \\
    & 5.20M & 8.89ms  & 92.61 $\pm$ 0.15 \\
    & 36.88M & 37.39ms  & \textbf{93.31 $\pm$ 0.05} \\
    \midrule
    \multirow{3}{*}{\parbox{2.3cm}{\centering NeuralScale (Iteration = 15)}}         
    & 0.58M & 2.94ms & \textbf{91.22 $\pm$ 0.15}     \\ 
    & 5.20M & 12.52ms  & \textbf{92.63 $\pm$ 0.12} \\               
    & 36.90M & 53.35ms  & 93.29 $\pm$ 0.09 \\ 
\midrule
    \multicolumn{4}{c}{\textbf{MobileNetV2 CIFAR100}} \\
    \multirow{3}{*}{\parbox{2.3cm}{\centering Uniform Scale (Baseline)}}   
    & 0.20M & 5.37ms & 57.80 $\pm$ 0.31     \\
    & 1.42M & 7.46ms  & 67.85 $\pm$ 0.38 \\
    & 9.30M & 19.69ms  & 72.40 $\pm$ 0.22 \\
    \midrule
    Li \etal \cite{li2016pruning}\tnote{$\dagger$} 
    & 1.42M & 7.71ms  & 67.12 $\pm$ 0.08 \\
    \midrule
    \multirow{3}{*}{\parbox{2.4cm}{\centering MorphNet \cite{gordon2018morphnet} (Taylor-FO \cite{molchanov2019importance})}}          
    & 0.20M & 6.14ms & 57.51 $\pm$ 0.36     \\
    & 1.42M & 10.95ms  & 67.51 $\pm$ 0.48 \\
    & 9.30M & 26.53ms  & 72.29 $\pm$ 0.28 \\
    \midrule
    \multirow{3}{*}{\parbox{2.3cm}{\centering NeuralScale (Iteration = 1)}}   
    & 0.19M & 5.69ms & 66.00 $\pm$ 0.12     \\
    & 1.40M & 11.73ms  & 70.23 $\pm$ 0.25 \\
    & 9.21M & 21.32ms  & 72.37 $\pm$ 0.12 \\
    \midrule
    \multirow{3}{*}{\parbox{2.3cm}{\centering NeuralScale (Iteration = 15)}}         
    & 0.19M & 7.84ms & \textbf{66.36 $\pm$ 0.28}     \\
    & 1.41M & 17.89ms  & \textbf{71.94 $\pm$ 0.45} \\
    & 9.27M & 40.48ms  & \textbf{74.73 $\pm$ 0.26} \\
    \midrule
    \multicolumn{4}{c}{\textbf{ResNet18 TinyImageNet}}\\
    \multirow{2}{*}{\parbox{2.3cm}{\centering Uniform Scale (Baseline)}}   
    & 0.73M & 3.02ms & 50.54 $\pm$ 0.37     \\
    & 6.36M & 11.56ms  & 56.68 $\pm$ 0.28 \\
    \midrule
    Li \etal \cite{li2016pruning}\tnote{$\dagger$} 
    & 6.36M & 11.93ms  & 54.72 $\pm$ 0.24 \\
    \midrule
    \multirow{2}{*}{\parbox{2.4cm}{\centering MorphNet \cite{gordon2018morphnet} (Taylor-FO \cite{molchanov2019importance})}}
    & 0.72M & 3.80ms & 50.79 $\pm$ 0.38     \\
    & 6.39M & 14.83ms  & 56.78 $\pm$ 0.85 \\
    \midrule
    \multirow{2}{*}{\parbox{2.3cm}{\centering NeuralScale (Iteration = 1)}}   
    & 0.72M & 5.96ms & 51.66 $\pm$ 0.80     \\
    & 6.42M & 14.58ms  & 57.89 $\pm$ 0.28 \\
    \midrule
    \multirow{2}{*}{\parbox{2.3cm}{\centering NeuralScale (Iteration = 15)}}         
    & 0.72M & 6.42ms & \textbf{53.95 $\pm$ 0.53}     \\
    & 6.40M & 17.52ms  & \textbf{58.40 $\pm$ 0.54} \\
\bottomrule
  \end{tabular}
  \begin{tablenotes}\footnotesize
\item[$\dagger$] Fine-tuned using pre-trained network (not trained from scratch).
\end{tablenotes}
\end{threeparttable}
\end{table}

\paragraph{MobileNetV2.}
The application to MobileNetV2 is to show the extensibility of our approach to a delicately handcrafted architecture. Our approach is superior in parameter efficiency (most cases) when compared to other methods as shown in Figure \ref{fig:mobilev2_c100} and \ref{fig:mobilev2_imagenet}. As shown in Table \ref{tab:acc_comparison}, an accuracy gain of 8.56\% for CIFAR100 relative to baseline is observed at a scaling ratio of 0.25. Our approach is also efficient in latency as shown in Figure \ref{fig:latency_mobilev2_c100}. An unintuitive observation can be seen on the experiment for TinyImageNet where the accuracy at iteration 1 outperforms iteration 15 for NeuralScale at ratio 0.25. The accuracy is below the baseline when more parameters are used. As the default configuration of MobileNetV2 has consistent filters, the default shortcut connections are identity mappings. We hypothesize that the switch from identity mapping to a convolutional mapping for the shortcut connection is the culprit. Empirical study is done in Supp.\ Section \ref{sec:convcut} with results supporting our hypothesis and explaining the observations.

\paragraph{ResNet18.}
As shown from the accuracy comparison plot in Figure \ref{fig:res18_c100} and \ref{fig:res18_imagenet}, substantial accuracy gain under different parameter scales is observed. From Table \ref{tab:acc_comparison}, an accuracy gain of 3.41\% is observed for TinyImageNet at a scale of 0.25. Accuracy gain using architecture descent is also more pronounced here. The accuracy gain here is in contradictory with the results in \cite{liu2018rethinking} (no gain in accuracy observed) probably due to the better pruning technique we use \cite{molchanov2019importance}. In Figure \ref{fig:latency_res18_c100} and \ref{fig:latency_res18_imagenet} our approach is comparable and in most cases better than the baseline configuration in latency.

\section{Conclusion}\label{sec:conclusion}
In this work, we propose a method to efficiently scale the neuron (filter) number of an architecture. We hypothesize that the scaling of network should follow a non-linear rule and is shown empirically that through our approach, networks efficient in parameters across different scales can be generated using this rule. Our empirical results on architecture descent also shed light on the efficient allocation of parameters in a deep neural network. As our approach is computationally-efficient and is complementary to most techniques and architectures, the inclusion to existing deep learning framework is cost-effective and results in a model of higher accuracy under the same parameter constraint.

\clearpage

{\small
\bibliographystyle{ieee_fullname}
\bibliography{egpaper_for_review}
}

\clearpage

\section{Supplementary Material}
\subsection{Update of $\tau$}\label{sec:update_tau}
To meet the memory limitation (parameter constraint) of any platform, we only need to update $\tau$ in our approach. We show the proof of the update of $\tau$ as follows:
\begin{align}
    \tau_i &= \tau_{i-1} - \Delta \tau_{i-1} \\
        &= \tau_{i-1} - \eta \frac{\partial \frac{1}{2} (h(f(\vec{x}|\vec{W},\vec{\Phi}(\tau_{i-1}|\vec{\Theta}))) - \hat{\tau}  )^2 }{\partial\tau_{i-1}} \\
        &= \tau_{i-1} - \eta (h(f(\vec{x}|\vec{W},\vec{\Phi}(\tau_{i-1}|\vec{\Theta}))) - \hat{\tau}  ) \\
        &\qquad \cdot\frac{\partial h(f(\vec{x}|\vec{W},\vec{\Phi}(\tau_{i-1}|\vec{\Theta})))}{\partial \vec{\Phi}(\tau_{i-1}|\vec{\Theta})}
        \frac{\partial \vec{\Phi}(\tau_{i-1}|\vec{\Theta})  }{\partial\tau_{i-1}} \label{eq:chain_rule} \\
        &= \tau_{i-1} - \eta(h(f(\vec{x}|\vec{W},\vec{\Phi}(\tau_{i-1}|\vec{\Theta}))) - \hat{\tau}  )\sum_{l=1}^L \beta_l \alpha_l \tau_{i-1}^{\beta_l - 1}.
\end{align}
Note that (\ref{eq:chain_rule}) comes from the chain rule of derivative.

\subsection{Accuracy Comparisons}\label{sec:acc_comparison}
In this section we show the tabularized comparison of VGG11-CIFAR100, MobileNetV2-CIFAR100 and ResNet18-TinyImageNet which was not shown in the main paper. As shown in Table \ref{tab:acc_comparison_remain}, an accuracy gain of 5.85\%, 2.40\% and 3.04\% is observed for VGG11, MobileNetV2 and ResNet18 on CIFAR100, TinyImageNet and CIFAR100 respectively.

\begin{table}[t]
\begin{threeparttable}
  \caption{Comparison of various network-dataset pairs.}
  \label{tab:acc_comparison_remain}
  \centering
  \begin{tabular}{cccc}
    \toprule
    Method & Params & Latency & Accuracy (\%) \\
    \midrule
    \midrule
    \multicolumn{4}{c}{\textbf{VGG11 CIFAR100}} \\
    \multirow{3}{*}{\parbox{2.3cm}{\centering Uniform Scale (Baseline)}}   
    & 0.59M & 1.30ms & 60.22 $\pm$ 0.45     \\
    & 5.23M & 4.28ms  & 68.56 $\pm$ 0.21 \\
    & 36.99M & 18.83ms  & 71.94 $\pm$ 0.25 \\
    \midrule
    Li \etal \cite{li2016pruning}\tnote{$\dagger$}         
    & 5.23M & 4.77ms & 68.41 $\pm$ 0.09     \\
    \midrule
    \multirow{3}{*}{\parbox{2.4cm}{\centering MorphNet \cite{gordon2018morphnet} (Taylor-FO \cite{molchanov2019importance})}}
    & 0.59M & 1.78ms & 64.85 $\pm$ 0.17     \\
    & 5.21M & 7.18ms  & 70.64 $\pm$ 0.38 \\
    & 36.80M & 41.52ms  & 72.72 $\pm$ 0.09 \\
    \midrule
    \multirow{3}{*}{\parbox{2.3cm}{\centering NeuralScale (Iteration = 1)}}         
    & 0.59M & 1.95ms & 65.71 $\pm$ 0.28     \\
    & 5.23M & 7.36ms  & 70.50 $\pm$ 0.16 \\
    & 36.98M & 33.24ms  & \textbf{72.78 $\pm$ 0.19} \\
    \midrule
    \multirow{3}{*}{\parbox{2.3cm}{\centering NeuralScale (Iteration = 15)}}         
    & 0.59M & 2.52ms & \textbf{66.07 $\pm$ 0.21}     \\
    & 5.23M & 10.19ms  & \textbf{70.70 $\pm$ 0.45} \\
    & 36.98M & 43.95ms  & \textbf{72.78 $\pm$ 0.13} \\              
 \midrule
\multicolumn{4}{c}{\textbf{MobileNetV2 TinyImageNet}} \\
    \multirow{2}{*}{\parbox{2.3cm}{\centering Uniform Scale (Baseline)}}   
    & 0.23 & 8.53ms & 44.22 $\pm$ 0.40     \\
    & 1.52M & 18.87ms  & 54.63 $\pm$ 0.46 \\
    \midrule
    Li \etal \cite{li2016pruning}\tnote{$\dagger$} 
    & 1.52M & 18.76ms  & 52.71 $\pm$ 0.28 \\
    \midrule
    \multirow{2}{*}{\parbox{2.4cm}{\centering MorphNet \cite{gordon2018morphnet} (Taylor-FO \cite{molchanov2019importance})}}
    & 0.23M & 10.47ms & 44.53 $\pm$ 0.50     \\
    & 1.51M & 28.88ms  & 53.08 $\pm$ 0.52 \\
    \midrule
    \multirow{2}{*}{\parbox{2.3cm}{\centering NeuralScale (Iteration = 1)}}   
    & 0.22M & 14.96ms & \textbf{49.70 $\pm$ 0.73}     \\
    & 1.49M & 26.98ms  & 54.18 $\pm$ 0.57 \\
    \midrule
    \multirow{2}{*}{\parbox{2.3cm}{\centering NeuralScale (Iteration = 15)}}         
    & 0.22M & 17.16ms & 46.82 $\pm$ 0.89     \\
    & 1.49M & 41.20ms  & \textbf{55.42 $\pm$ 0.44} \\
    \midrule
    \multicolumn{4}{c}{\textbf{ResNet18 CIFAR100}} \\
    \multirow{3}{*}{\parbox{2.3cm}{\centering Uniform Scale (Baseline)}}   
    & 0.71M & 2.53ms & 68.10 $\pm$ 0.40     \\
    & 6.32M & 9.98ms  & 75.10 $\pm$ 0.34 \\
    & 44.75M & 47.04ms  & 78.39 $\pm$ 0.29 \\
    \midrule
    Li \etal \cite{li2016pruning}\tnote{$\dagger$} 
    & 6.32M & 10.18ms  & 73.91 $\pm$ 0.12 \\
    \midrule
    \multirow{3}{*}{\parbox{2.4cm}{\centering MorphNet \cite{gordon2018morphnet} (Taylor-FO \cite{molchanov2019importance})}}
    & 0.72M & 3.73ms & 69.34 $\pm$ 0.31     \\
    & 6.29M & 15.03ms  & 75.60 $\pm$ 0.40 \\
    & 44.53M & 98.54ms  & \textbf{78.68 $\pm$ 0.17} \\
    \midrule
    \multirow{3}{*}{\parbox{2.3cm}{\centering NeuralScale (Iteration = 1)}}         
    & 0.71M & 4.51ms & 70.63 $\pm$ 0.13     \\
    & 6.38M & 11.95ms  & 75.83 $\pm$ 0.15 \\
    & 45.15M & 50.24ms  & 78.39 $\pm$ 0.22 \\
    \midrule
    \multirow{3}{*}{\parbox{2.3cm}{\centering NeuralScale (Iteration = 15)}}         
    & 0.71M & 5.71ms & \textbf{71.14 $\pm$ 0.45}     \\
    & 6.36M & 19.54ms  & \textbf{76.35 $\pm$ 0.20} \\
    & 45.05M & 90.18ms  & 78.62 $\pm$ 0.13 \\
\bottomrule
  \end{tabular}
  \begin{tablenotes}\footnotesize
\item[$\dagger$] Fine-tuned using pre-trained network (not trained from scratch).
\end{tablenotes}
\end{threeparttable}
\end{table}

\subsection{Pre-Training Epochs $P$}\label{sec:pre-training}
The pruning of neural network is usually done on a pre-trained network. As we want our algorithm to be efficient in terms of search cost, we explore the possibility of reduction in time or epochs for network pre-training by tuning the pre-training epochs $P$. To our surprise, having a large $P$ does not result in an architecture with the best performance. Here, we investigate how $P$ affects the accuracy of the final configuration, proving that conventional wisdom on when to apply pruning might be flawed. Experiments will be shown on VGG11 and MobileNetV2 on CIFAR10 and CIFAR100 respectively. All results shown are based on the final (iteration=15) iteration of architecture descent.

\begin{figure*}[!tbp]
\centering
  	\subfloat[$P=0$]{\includegraphics[width=0.5\textwidth]{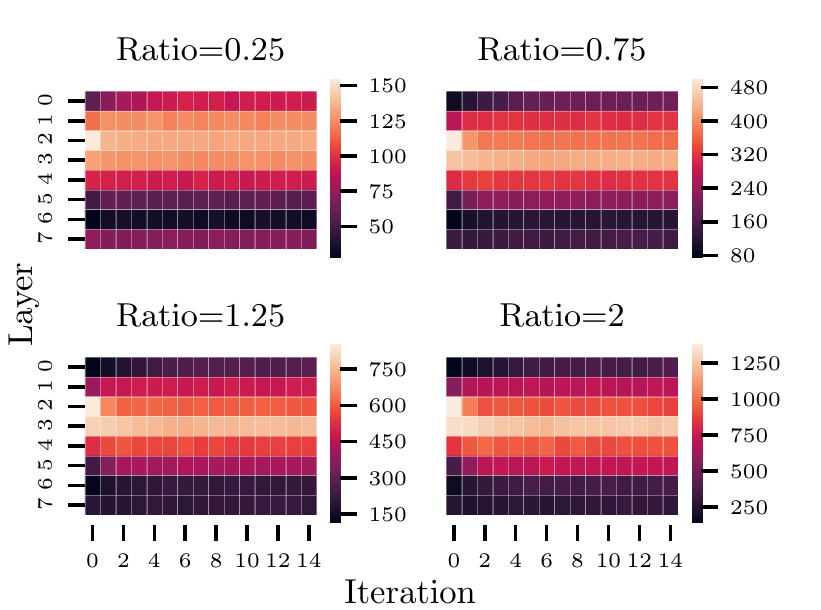}\label{fig:vgg_p0}}
	\hfill
  	\subfloat[$P=2$]{\includegraphics[width=0.5\textwidth]{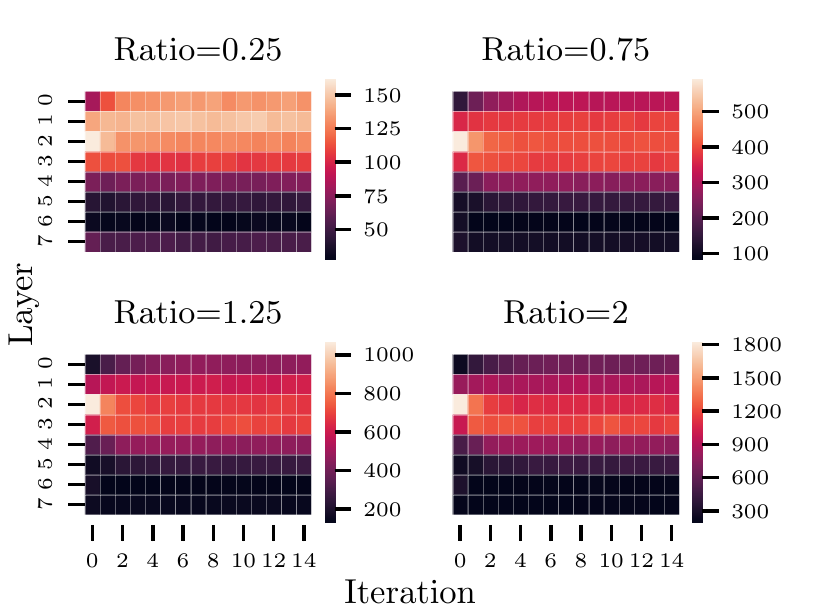}\label{fig:vgg_p2}}
	\vfill
  	\subfloat[$P=5$]{\includegraphics[width=0.5\textwidth]{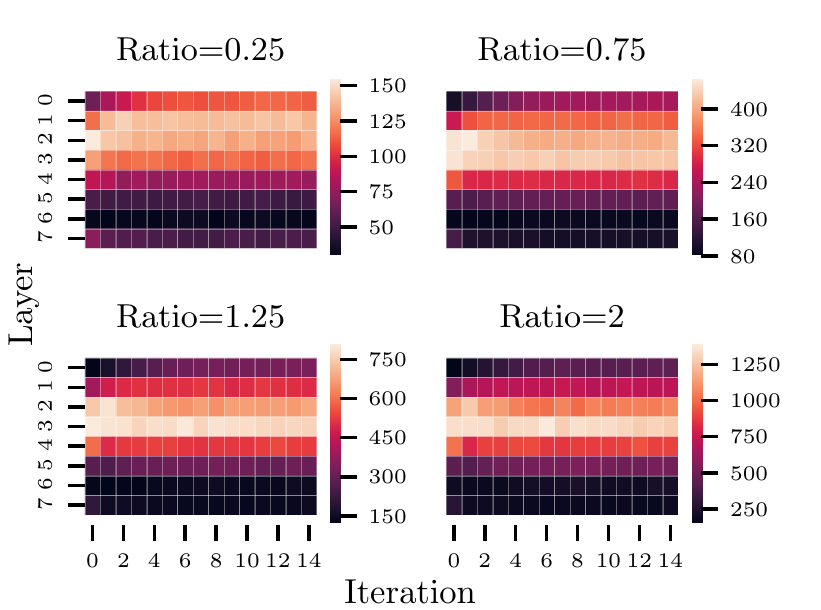}\label{fig:vgg_p5}}
	\hfill
  	\subfloat[$P=10$]{\includegraphics[width=0.5\textwidth]{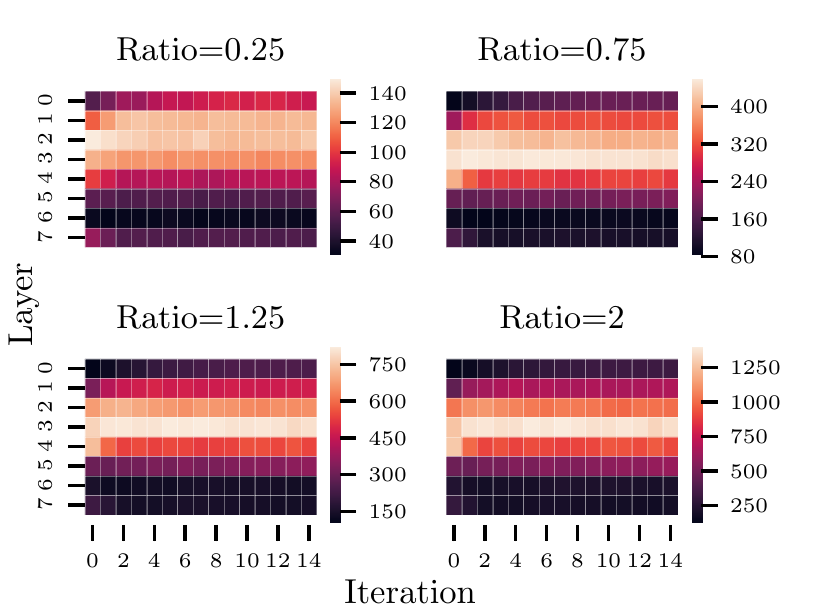}\label{fig:vgg_p10}}
	\vfill
  	\subfloat[$P=30$]{\includegraphics[width=0.5\textwidth]{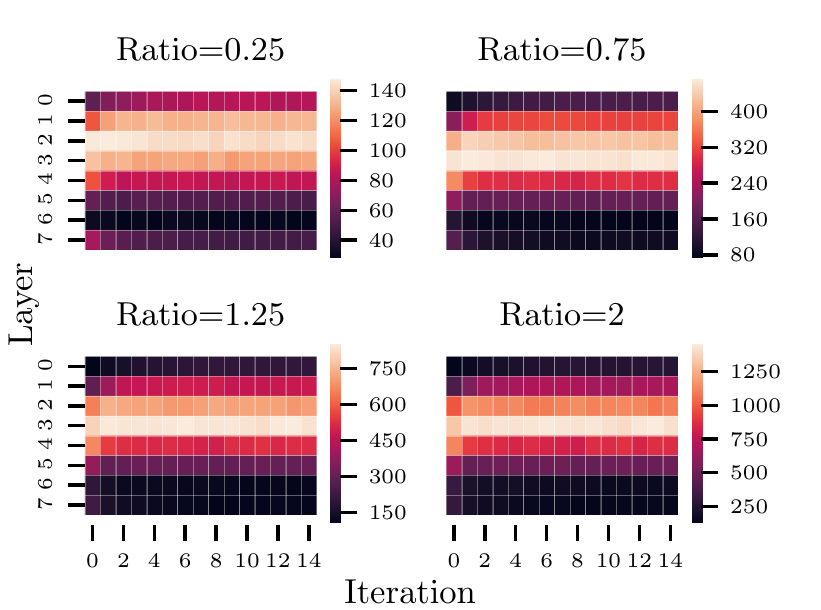}\label{fig:vgg_p30}}
	\hfill
  	\subfloat[$P=60$]{\includegraphics[width=0.5\textwidth]{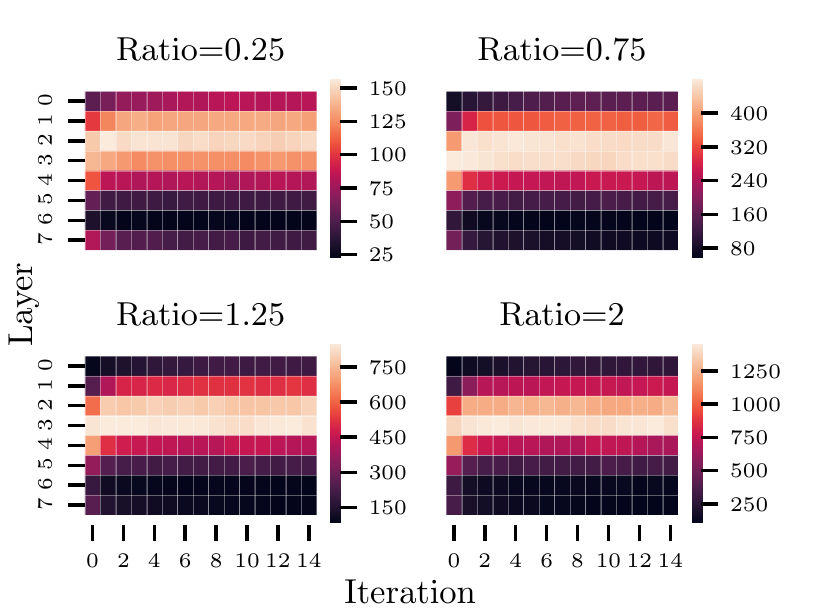}\label{fig:vgg_p60}}
\caption{Showing the difference in searched architecture by running architecture descent on VGG11 for CIFAR10 using various value of pre-training epochs $P$.}
\label{fig:architecuture_descent_vgg11_pre-training}
\end{figure*}

\begin{figure}
    \centering
    \subfloat[Accuracy vs Parameter.]{\includegraphics[width=\columnwidth]{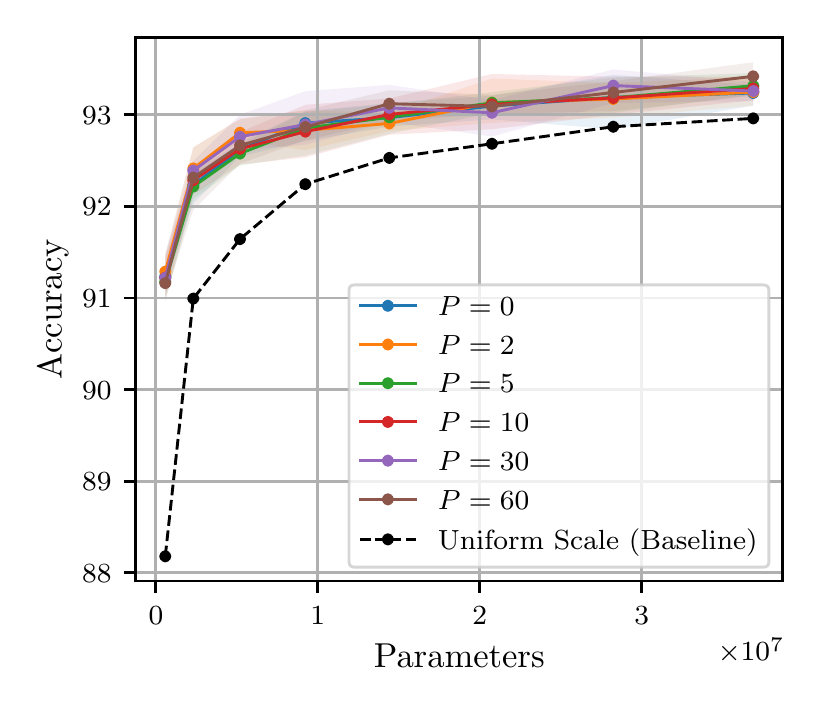}\label{fig:vgg11_pre-training_param}}
	\vfill
	\subfloat[Accuracy vs Latency.]{\includegraphics[width=\columnwidth]{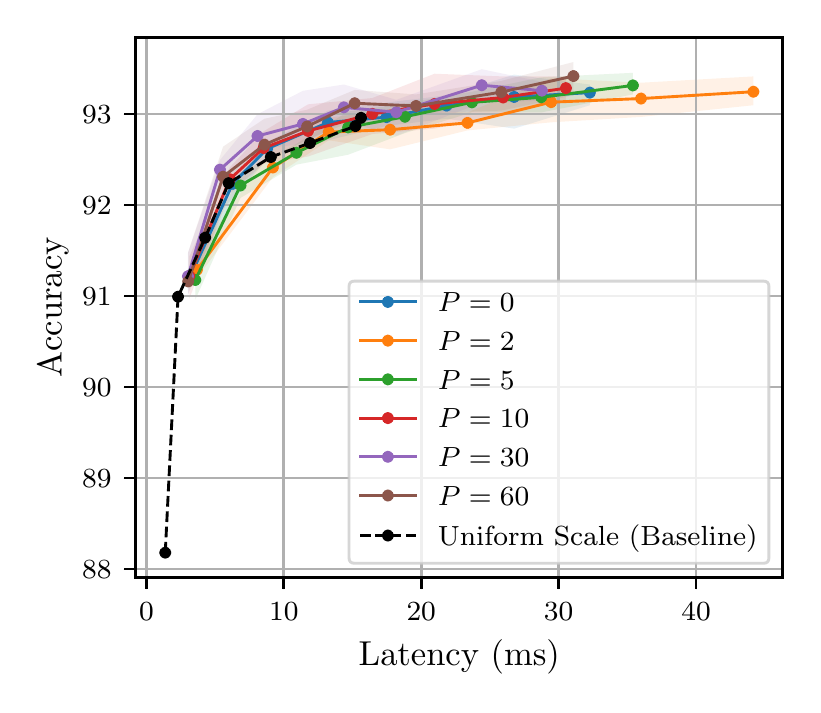}\label{fig:vgg11_pre-training_latency}}
    \caption{Accuracy comparison plot for VGG11 on CIFAR10 that uses different pre-training epochs $P$ before pruning. \protect\subref{fig:vgg11_pre-training_param} shows the accuracy comparison under different parameters using different value of $P$. \protect\subref{fig:vgg11_pre-training_latency} shows the comparison of accuracy under different latencies using different value of $P$.}
    \label{fig:vgg11_pre-training}
\end{figure}

\paragraph{VGG11.}
CIFAR10 will be used for the experimentation on $P$ for VGG11. We show architecture and results obtained by setting $P$ to be 0, 2, 5, 10, 30 and 60. The searched architecture is shown in Figure \ref{fig:architecuture_descent_vgg11_pre-training}. We next show the comparison plot using different pre-training epochs in Figure \ref{fig:vgg11_pre-training} accompanied by Table \ref{tab:vgg_pre-training}. For a simple network like VGG11, the number of pre-training epochs doesn't have too much of an impact in performance which can be clearly observed in the resulting filter configuration in Figure \ref{fig:architecuture_descent_vgg11_pre-training}.

\begin{table}
  \caption{Accuracy comparison on VGG11 for CIFAR10 using different pre-training epochs $P$.}
  \label{tab:vgg_pre-training}
  \centering
  \begin{tabular}{cccc}
    \toprule
    Method & Params & Latency & Accuracy (\%) \\
    \midrule
    \midrule
    \multirow{3}{*}{\parbox{2.3cm}{\centering Uniform Scale (Baseline)}}   
    & 0.58M & 1.30ms & 88.18 $\pm$ 0.16     \\
    & 5.20M & 4.31ms  & 91.64 $\pm$ 0.10 \\
    & 36.89M & 19.50ms  & 92.96 $\pm$ 0.09 \\
    \midrule
    \multirow{3}{*}{\parbox{2.3cm}{\centering NeuralScale ($P=0$)}}   
    & 0.58M & 3.01ms & 91.23 $\pm$ 0.05     \\
    & 5.20M & 12.35ms  & 92.62 $\pm$ 0.06 \\
    & 36.89M & 53.26ms  & 93.24 $\pm$ 0.09 \\
    \midrule
    \multirow{3}{*}{\parbox{2.3cm}{\centering NeuralScale ($P=2$)}}   
    & 0.58M & 3.49ms & \textbf{91.29 $\pm$ 0.09}     \\
    & 5.20M & 20.24ms  & \textbf{92.80 $\pm$ 0.09} \\
    & 36.90M & 81.27ms  & 93.25 $\pm$ 0.10 \\
    \midrule
    \multirow{3}{*}{\parbox{2.3cm}{\centering NeuralScale ($P=5$)}}   
    & 0.58M & 3.34ms & 91.18 $\pm$ 0.13     \\
    & 5.19M & 17.30ms  & 92.58 $\pm$ 0.08 \\
    & 36.90M & 63.85ms  & 93.31 $\pm$ 0.08 \\
    \midrule
    \multirow{3}{*}{\parbox{2.3cm}{\centering NeuralScale ($P=10$)}}   
    & 0.58M & 2.93ms & 91.22 $\pm$ 0.15     \\
    & 5.20M & 12.53ms  & 92.63 $\pm$ 0.12 \\
    & 36.90M & 55.44ms  & 93.29 $\pm$ 0.09 \\
    \midrule
    \multirow{3}{*}{\parbox{2.3cm}{\centering NeuralScale ($P=30$)}}   
    & 0.58M & 2.82ms & 91.22 $\pm$ 0.15     \\
    & 5.20M & 11.85ms  & 92.76 $\pm$ 0.13 \\
    & 36.90M & 51.02ms  & 93.26 $\pm$ 0.08 \\
    \midrule
    \multirow{3}{*}{\parbox{2.3cm}{\centering NeuralScale ($P=60$)}}   
    & 0.58M & 2.85ms & 91.16 $\pm$ 0.17     \\
    & 5.20M & 12.58ms  & 92.66 $\pm$ 0.18 \\
    & 36.89M & 61.14ms  & \textbf{93.42 $\pm$ 0.13} \\
    \bottomrule
  \end{tabular}
\end{table}

\begin{figure*}[!tbp]
\centering
  	\subfloat[$P=0$]{\includegraphics[width=0.5\textwidth]{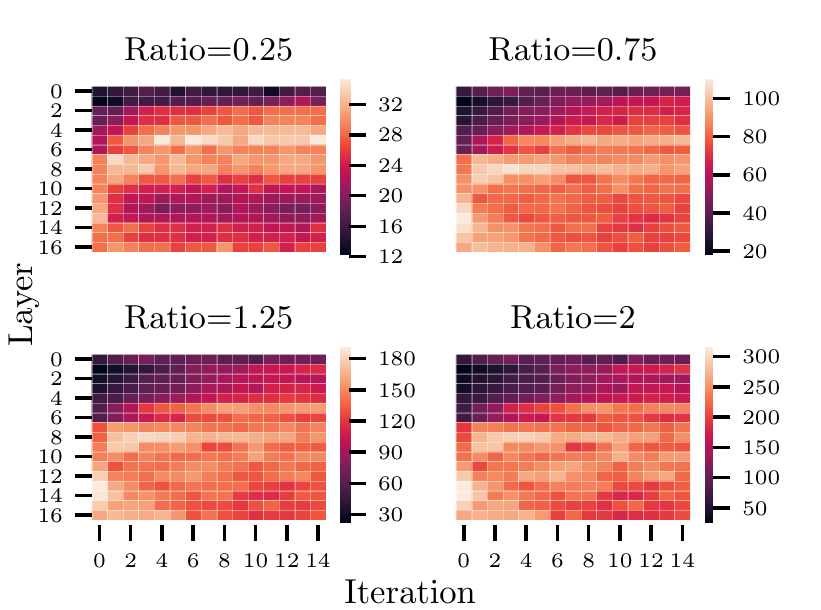}\label{fig:mobile_p0}}
	\hfill
  	\subfloat[$P=2$]{\includegraphics[width=0.5\textwidth]{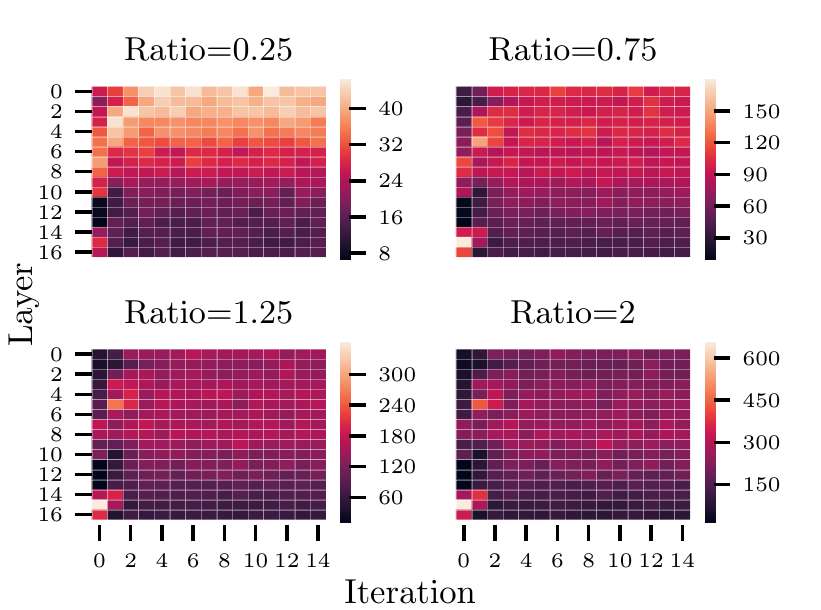}\label{fig:mobile_p2}}
	\vfill
  	\subfloat[$P=5$]{\includegraphics[width=0.5\textwidth]{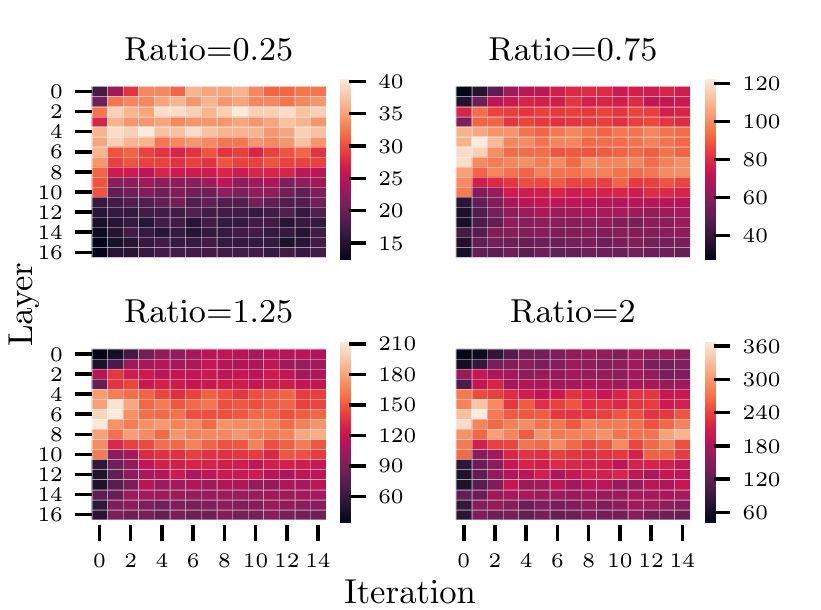}\label{fig:mobile_p5}}
	\hfill
  	\subfloat[$P=10$]{\includegraphics[width=0.5\textwidth]{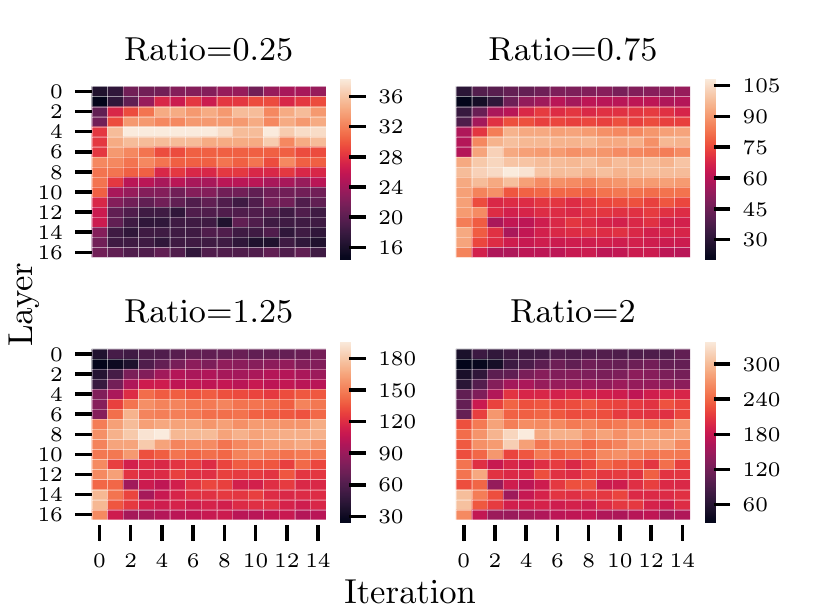}\label{fig:mobile_p10}}
	\vfill
  	\subfloat[$P=30$]{\includegraphics[width=0.5\textwidth]{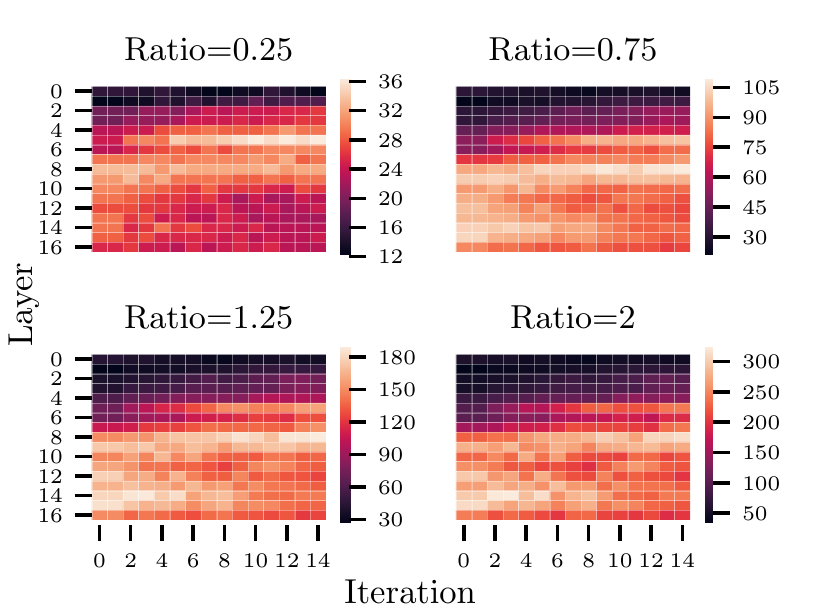}\label{fig:mobile_p30}}
	\hfill
  	\subfloat[$P=60$]{\includegraphics[width=0.5\textwidth]{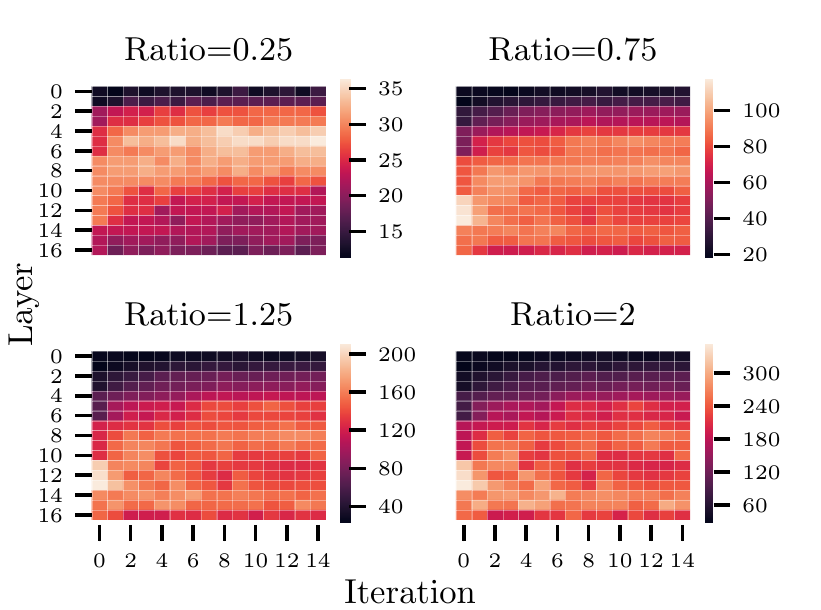}\label{fig:mobile_p60}}
\caption{Showing the difference in searched architecture by running architecture descent on MobileNetV2 for CIFAR100 using various value of pre-training epochs $P$.}
\label{fig:architecuture_descent_mobilenetv2_pre-training}
\end{figure*}

\begin{figure}
    \centering
    \subfloat[Accuracy vs Parameters.]{\includegraphics[width=\columnwidth]{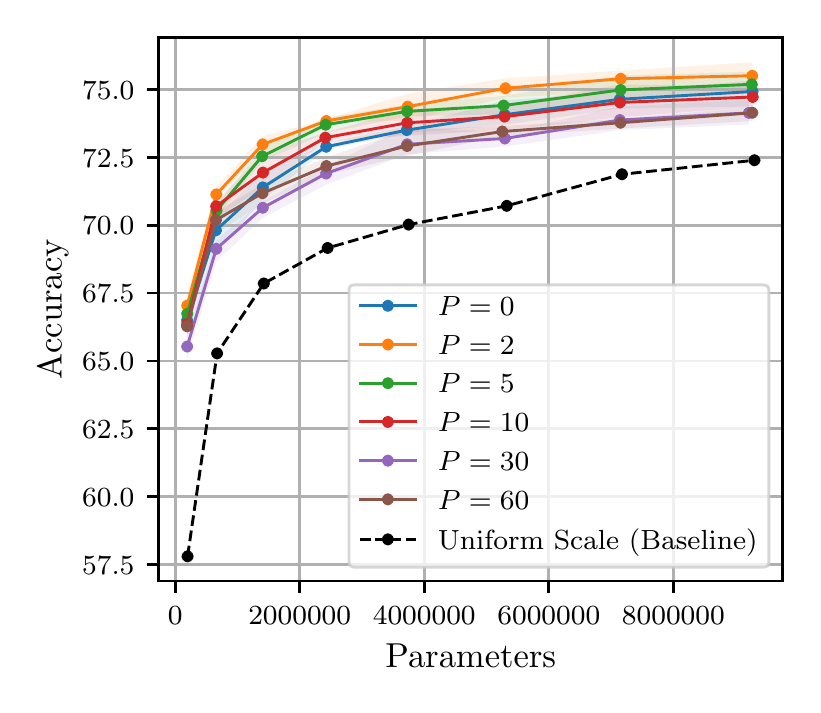}\label{fig:mobilenetv2_pre-training_param}}
	\vfill
	\subfloat[Accuracy vs Latency.]{\includegraphics[width=\columnwidth]{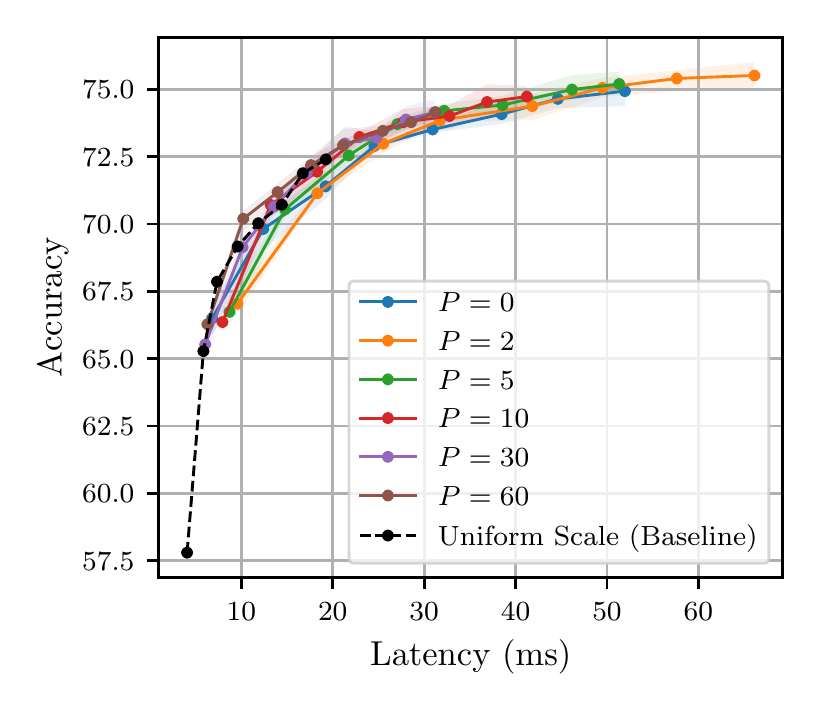}\label{fig:mobilenetv2_pre-training_latency}}
    \caption{Accuracy comparison plot for MobileNetV2 on CIFAR100 that uses different pre-training epochs $P$ before pruning. \protect\subref{fig:mobilenetv2_pre-training_param} shows the accuracy comparison under different parameters using different value of $P$. \protect\subref{fig:mobilenetv2_pre-training_latency} shows the comparison of accuracy under different latencies using different value of $P$.}
    \label{fig:mobilenetv2_pre-training}
\end{figure}

\paragraph{MobileNetV2.}
CIFAR100 will be used for the experimentation on $P$ for MobileNetV2. We show architecture and results obtained by setting $P$ to be 0, 2, 5, 10, 30 and 60. The searched architecture is shown in Figure \ref{fig:architecuture_descent_mobilenetv2_pre-training}. We next show the comparison plot using different pre-training epochs in Figure \ref{fig:mobilenetv2_pre-training} accompanied by Table \ref{tab:mobilenetv2_pre-training}. It is interesting to see that for a deeper and more complicated network like MobileNetV2, there's a notable variation in the distribution of filters with respect to the number of pre-training epochs. The accuracy comparison in Figure \ref{fig:mobilenetv2_pre-training} shows that having large number of pre-training epochs doesn't help the efficiency in parameters and instead impedes it. It is shown that $P=2$ or $P=5$ gives us a configuration of filters that is the most efficient in terms of parameters for MobileNetV2 on CIFAR100. This is an interesting observation which sheds light on the number of pre-training iterations required prior to network pruning for optimal performance.

\begin{table}
  \caption{Accuracy comparison on MobileNetV2 for CIFAR100 using different pre-training epochs $P$.}
  \label{tab:mobilenetv2_pre-training}
  \centering
  \begin{tabular}{cccc}
    \toprule
    Method & Params & Latency & Accuracy (\%) \\
    \midrule
    \midrule
    \multirow{3}{*}{\parbox{2.3cm}{\centering Uniform Scale (Baseline)}}   
    & 0.20M & 5.53ms & 57.80 $\pm$ 0.31     \\
    & 1.42M & 7.56ms  & 67.85 $\pm$ 0.38 \\
    & 9.30M & 20.43ms  & 72.40 $\pm$ 0.22 \\
    \midrule
    \multirow{3}{*}{\parbox{2.3cm}{\centering NeuralScale ($P=0$)}}   
    & 0.19M & 6.64ms & 66.49 $\pm$ 0.43     \\
    & 1.40M & 18.77ms  & 71.39 $\pm$ 0.45 \\
    & 9.27M & 52.49ms  & 74.93 $\pm$ 0.34 \\
    \midrule
    \multirow{3}{*}{\parbox{2.3cm}{\centering NeuralScale ($P=2$)}}   
    & 0.19M & 9.42ms & \textbf{67.04 $\pm$ 0.28}     \\
    & 1.40M & 24.95ms  & \textbf{72.98 $\pm$ 0.26} \\
    & 9.27M & 67.55ms  & \textbf{75.51 $\pm$ 0.41} \\
    \midrule
    \multirow{3}{*}{\parbox{2.3cm}{\centering NeuralScale ($P=5$)}}   
    & 0.19M & 8.61ms & 66.74 $\pm$ 0.39     \\
    & 1.40M & 21.37ms  & 72.54 $\pm$ 0.18 \\
    & 9.26M & 51.63ms  & 75.19 $\pm$ 0.26 \\
    \midrule
    \multirow{3}{*}{\parbox{2.3cm}{\centering NeuralScale ($P=10$)}}   
    & 0.19M & 7.82ms & 66.36 $\pm$ 0.28     \\
    & 1.41M & 18.02ms  & 71.94 $\pm$ 0.45 \\
    & 9.27M & 43.00ms  & 74.73 $\pm$ 0.26 \\
    \midrule
    \multirow{3}{*}{\parbox{2.3cm}{\centering NeuralScale ($P=30$)}}   
    & 0.19M & 6.20ms & 65.53 $\pm$ 0.31     \\
    & 1.41M & 13.35ms  & 70.64 $\pm$ 0.23 \\
    & 9.21M & 31.77ms  & 74.14 $\pm$ 0.35 \\
    \midrule
    \multirow{3}{*}{\parbox{2.3cm}{\centering NeuralScale ($P=60$)}}   
    & 0.19M & 6.15ms & 66.28 $\pm$ 0.13     \\
    & 1.40M & 13.74ms  & 71.18 $\pm$ 0.24 \\
    & 9.27M & 32.15ms  & 74.15 $\pm$ 0.18 \\
    \bottomrule
  \end{tabular}
\end{table}

\subsection{Using Convolutional Layers as Shortcut Connection}\label{sec:convcut}
\begin{figure}
    \centering
    \subfloat[Accuracy vs Parameter.]{\includegraphics[width=\columnwidth]{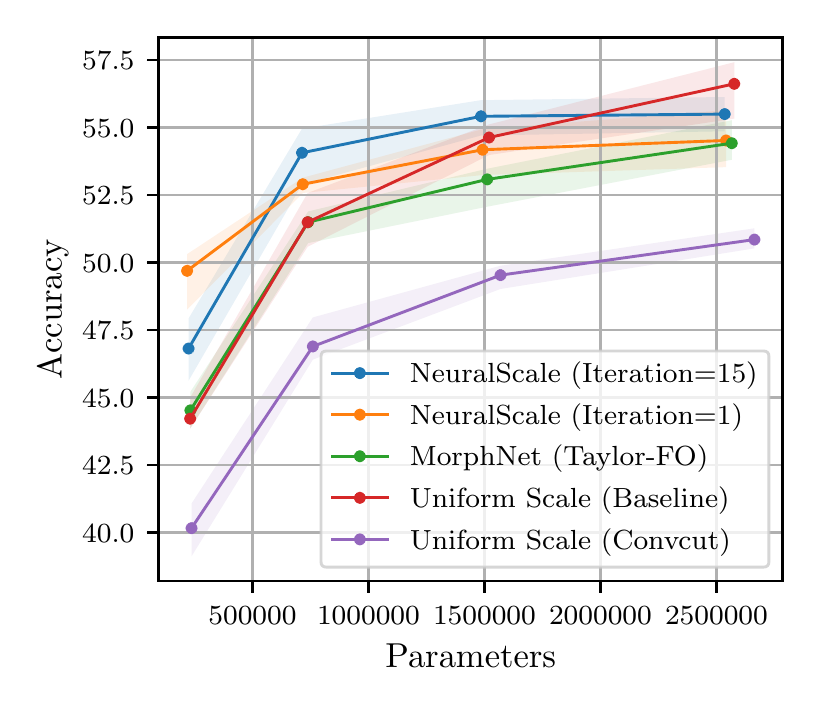}\label{fig:convcut_param_mobilenetv2}}
	\vfill
	\subfloat[Accuracy vs Latency.]{\includegraphics[width=\columnwidth]{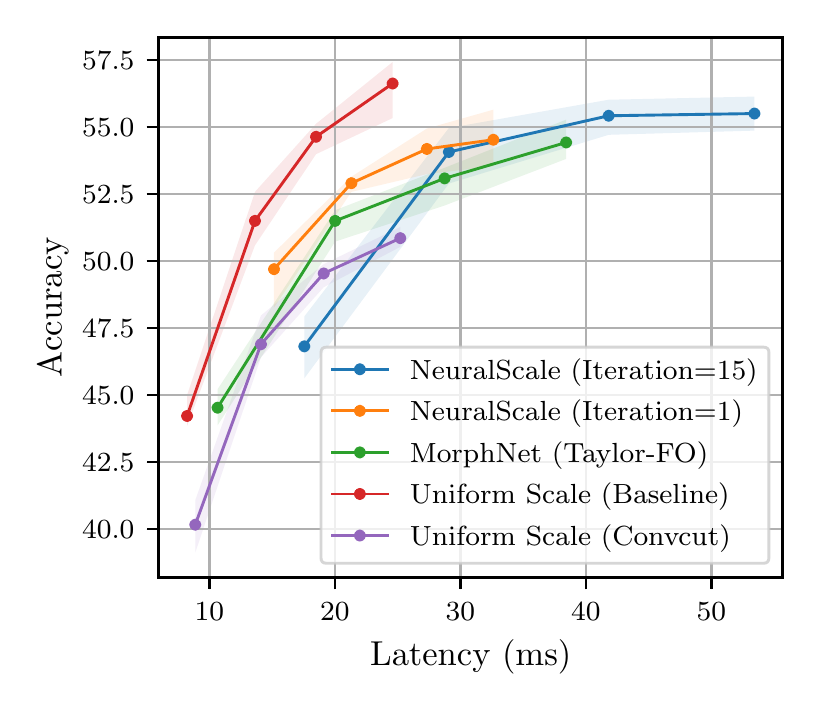}\label{fig:convcut_latency_mobilenetv2}}
    \caption{Accuracy comparison plot for MobileNetV2 on TinyImageNet with inclusion of ConvCut.}
    \label{fig:convcut_mobilenetv2}
\end{figure}

\begin{figure}
    \centering
    \subfloat[Accuracy vs Parameter.]{\includegraphics[width=\columnwidth]{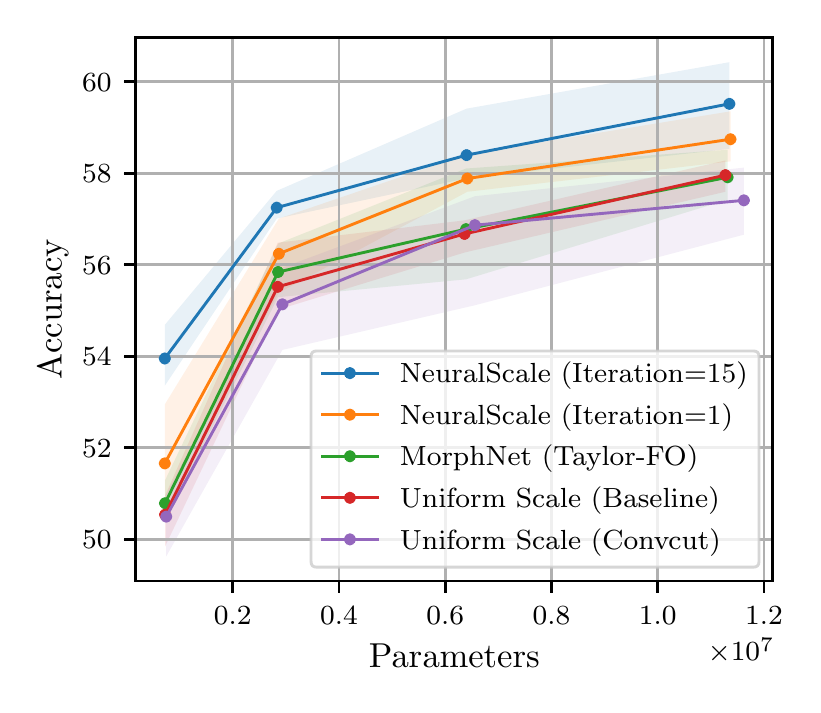}\label{fig:convcut_param_resnet18}}
	\vfill
	\subfloat[Accuracy vs Latency.]{\includegraphics[width=\columnwidth]{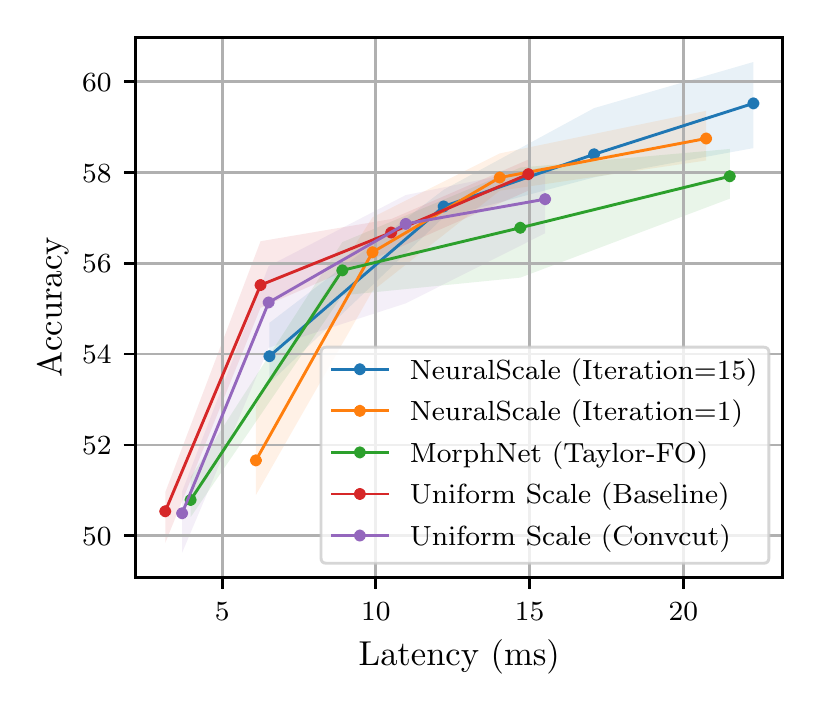}\label{fig:convcut_latency_resnet18}}
    \caption{Accuracy comparison plot for ResNet18 on TinyImageNet with inclusion of ConvCut.}
    \label{fig:convcut_resnet18}
\end{figure}

By default, MobileNetV2 has shortcut connections composed of identity mappings. By modifying the filter sizes of MobileNetV2, the shortcut connection has to be changed to a convolutional one instead to compensate the difference in filter sizes on both ends of the shortcut connection. A surprising finding is that the change from identity mapping to convolutional mapping affects the original performance significantly, despite the increase in parameter. We show experiments comparing two kinds of shortcut connection (identity and convolutional) on the original configuration which is uniformly scaled to different ratios. We name the method that uses convolutional shortcuts as \textit{ConvCut}. A comparison plot comparing ConvCut with other scaling methods using ResNet18 and MobileNetV2 on TinyImageNet is shown in Figure \ref{fig:convcut_mobilenetv2} and \ref{fig:convcut_resnet18} respectively. Results are summarized in Table \ref{tab:convcut_resnet18} and Table \ref{tab:convcut_mobilenetv2} for ResNet18 and MobileNetV2 respectively. It can be observed that the switch from identity to convolutional mapping doesn't have drastic impact on the accuracy of ResNet18 but a significant drop in accuracy can be observed for MobileNetV2. Our conjecture is that the design of linear bottleneck layers in MobileNetV2 is to embed a low-dimensional manifold where switching from identity to convolutional mapping for shortcut layer that connects linear bottleneck layers introduces noise to this manifold which is harmful for information propagation and network training. Despite from the setback of accuracy drop through the introduction of convolutional shortcut layers, our approach is still able to induce accuracy gain in a low parameter count setting when compared to the baseline configuration setting, showing the importance of searching for the optimal configuration of filters. An unbiased comparison is to compare our approach with the convolutional shortcut (ConvCut) version of MobileNetV2 using the default set of filter configuration as shown in Figure \ref{fig:convcut_mobilenetv2} where both (ours and ConvCut) use convolutional layer as shortcut connection. On an apple-to-apple comparison, our approach shows superiority in parameter efficiency. This empirical study also explains the superiority in accuracy of iteration 1 when compared to iteration 15 of our approach as can be observed in Figure \ref{fig:convcut_param_mobilenetv2}. From our observation, iteration 1 of our approach generates a configuration composed repeated filters on some blocks, resulting in an architecture consisting of both identity and convolutional shortcut connection. Hence, it is not surprising that iteration 1 outperforms iteration 15 of our approach as it has both traits: identity shortcut and optimized filter configuration.

\begin{table}[t]
  \caption{Accuracy comparison on MobileNetV2 on TinyImageNet (includes ConvCut).}
  \label{tab:convcut_mobilenetv2}
  \centering
  \begin{tabular}{cccc}
    \toprule
    Method & Params & Latency & Accuracy (\%) \\
    \midrule
    \midrule
    \multirow{3}{*}{\parbox{2.3cm}{\centering Uniform Scale (Baseline)}}   
    & 0.23M & 8.53ms & 44.22 $\pm$ 0.40     \\
    & 1.52M & 18.87ms  & 54.63 $\pm$ 0.46 \\
    & 2.58M & 24.70ms  & \textbf{56.62 $\pm$ 0.70} \\
    \midrule
    \multirow{3}{*}{\parbox{2.4cm}{\centering MorphNet \cite{gordon2018morphnet} (Taylor-FO \cite{molchanov2019importance})}}
    & 0.23M & 10.47ms & 44.53 $\pm$ 0.50     \\
    & 1.51M & 28.88ms  & 53.08 $\pm$ 0.52 \\
    & 2.57M & 38.45ms  & 54.42 $\pm$ 0.53 \\
    \midrule
    \multirow{3}{*}{\parbox{2.3cm}{\centering Uniform Scale (ConvCut)}}         
    & 0.24M & 9.23ms & 40.16 $\pm$ 0.63     \\
    & 1.57M & 19.23ms  & 49.54 $\pm$ 0.30 \\
    & 2.66M & 25.39ms  & 50.85 $\pm$ 0.27 \\
    \midrule
    \multirow{3}{*}{\parbox{2.3cm}{\centering NeuralScale (Iteration = 1)}}         
    & 0.22M & 14.96ms & \textbf{49.70 $\pm$ 0.73}     \\
    & 1.49M & 26.98ms  & 54.18 $\pm$ 0.57 \\
    & 2.54M & 32.09ms  & 54.52 $\pm$ 0.72 \\
    \midrule
    \multirow{3}{*}{\parbox{2.3cm}{\centering NeuralScale (Iteration = 15)}}         
    & 0.22M & 17.16ms & 46.82 $\pm$ 0.89     \\
    & 1.49M & 41.20ms  & \textbf{55.42 $\pm$ 0.44} \\
    & 2.54M & 52.76ms  & 55.50 $\pm$ 0.51 \\
\bottomrule
  \end{tabular}
\end{table}

\begin{table}[t]
  \caption{Accuracy comparison on ResNet18 on TinyImageNet (includes ConvCut).}
  \label{tab:convcut_resnet18}
  \centering
  \begin{tabular}{cccc}
    \toprule
    Method & Params & Latency & Accuracy (\%) \\
    \midrule
    \midrule
    \multirow{3}{*}{\parbox{2.3cm}{\centering Uniform Scale (Baseline)}}   
    & 0.73M & 3.02ms & 50.54 $\pm$ 0.37     \\
    & 6.36M & 11.56ms  & 56.68 $\pm$ 0.28 \\
    & 11.27M & 15.46ms  & 57.96 $\pm$ 0.23 \\
    \midrule
    \multirow{3}{*}{\parbox{2.4cm}{\centering MorphNet \cite{gordon2018morphnet} (Taylor-FO \cite{molchanov2019importance})}}           
    & 0.72M & 3.80ms & 50.79 $\pm$ 0.38     \\
    & 6.39M & 14.83ms  & 56.78 $\pm$ 0.85 \\
    & 11.31M & 22.07ms  & 57.91 $\pm$ 0.38 \\
    \midrule
    \multirow{3}{*}{\parbox{2.3cm}{\centering Uniform Scale (ConvCut)}}         
    & 0.75M & 3.64ms & 50.50 $\pm$ 0.46     \\
    & 6.56M & 11.99ms  & 56.87 $\pm$ 0.88 \\
    & 11.62M & 15.98ms  & 57.41 $\pm$ 0.58 \\
    \midrule
    \multirow{3}{*}{\parbox{2.3cm}{\centering NeuralScale (Iteration = 1)}}         
    & 0.72M & 5.96ms & 51.66 $\pm$ 0.80     \\
    & 6.42M & 14.58ms  & 57.89 $\pm$ 0.28 \\
    & 11.37M & 22.11ms  & 58.75 $\pm$ 0.37 \\
    \midrule
    \multirow{3}{*}{\parbox{2.3cm}{\centering NeuralScale (Iteration = 15)}}         
    & 0.72M & 6.42ms & \textbf{53.95 $\pm$ 0.53}     \\
    & 6.40M & 17.52ms  & \textbf{58.40 $\pm$ 0.54} \\
    & 11.35M & 25.94ms  & \textbf{59.52 $\pm$ 0.63} \\
\bottomrule
  \end{tabular}
\end{table}

\end{document}